**PepEVOLVE: Position-Aware Dynamic Peptide Optimization via Group-Relative Advantage**

Trieu Nguyen[1,3], Peter Zhiping Zhang[1], Nicolas Boyer[1], Cheng Fang[2], Liying Zhang[1], Sebastian Schneider[1], Hao-Wei Pang[2*], Shasha Feng[1*]

1. Merck & Co., Inc., Boston, MA, USA, 02115
2. Merck & Co., Inc., Cambridge, MA, USA, 02141
3. University of South Florida, Tampa, FL, USA, 33620

*Corresponding authors (hao-wei.pang@merck.com, shasha.feng@merck.com)

## Abstract

Macrocyclic peptides are an emerging modality that combines biologics-like affinity with small-molecule-like developability, yet their vast combinatorial space and multi-parameter objectives make lead optimization slow and challenging. Prior generative approaches for peptide lead optimization (e.g., PepINVENT) require chemists to pre-specify mutable positions for optimization, choices that are not always known *a priori*, and rely on static pretraining and optimization algorithms that limit the model's ability to generalize and effectively optimize peptide sequences. We introduce PepEVOLVE, a position-aware, dynamic framework that learns both where to edit and how to dynamically optimize peptides for multi-parameter improvement. PepEVOLVE (i) augments pretraining with dynamic masking and CHUCKLES shifting to improve generalization, (ii) uses a context-free multi-armed-bandit router that discovers high-reward residues, and (iii) couples a novel evolving optimization algorithm with group-relative advantage to stabilize reinforcement updates. During *in silico* evaluations, the router policy reliably learns and concentrates probability on chemically tractable sites that influence the desired peptide's properties. On a therapeutically motivated Rev-binding macrocycle benchmark, PepEVOLVE outperformed PepINVENT by reaching higher averaged permeability-weighted optimization scores of ~0.8 vs. ~0.6, achieving best candidates with score of 0.97 (vs. ~0.89), and converging in fewer steps under the task of optimizing permeability and lipophilicity with structure constraints. Overall, PepEVOLVE offers a practical, reproducible path to linear and macrocyclic peptide lead optimization when optimal edit sites are unknown, enabling more efficient chemical space exploration and improving design quality across multiple objectives to accelerate the design cycles.

## 1. Introduction

### 1.1 Challenges in Peptide Drug Discovery and Design

Macrocyclic peptides are cyclic peptides, typically comprising 4-25 amino acids, in which the termini and/or side chains are covalently linked to form a ring.[1, 2] This cyclized topology can confer improved proteolytic stability, enhanced affinity and selectivity, extended half-life, and the ability to engage shallow protein–protein interaction surfaces that are often inaccessible to small molecules, making macrocyclic peptides an attractive modality for challenging intra- and extracellular targets.[3] This is exemplified by several oral macrocyclic peptides currently approved or in clinical development, including MSD's PCSK9 inhibitor LIPFENDRA (enlicitide, MK-0616),[4] Chugai's pan-KRAS inhibitor paluratide (LUNA18),[5] BMS's PD-L1 inhibitor BMS-986238,[6] and Protagonist/J&J IL-23 oral peptide inhibitor ICOTYDE (icotrokinra, JNJ-77242113/PN-235).[7]

Macrocyclic peptide lead optimization presents several challenges that distinguish it from traditional small molecule drug discovery. First, recent approved or in-clinical macrocyclic peptides often contain ~11–18 residues, resulting in many positions that must be optimized simultaneously.[5, 8, 9] Second, the incorporation of non-canonical amino acids (NCAAs) dramatically expands the accessible chemical space through diverse side chains, backbone-modified residues (non-α amino acids), and *N*-alkylation, and this space continues to grow exponentially as new building blocks become commercially, synthetically, or biocatalytically available.[8] Third, peptide optimization is inherently multi-parametric: potency, solubility, permeability, metabolic stability, and pharmacokinetics must be balanced within a highly non-linear, high-dimensional landscape, where improvements in one property can adversely affect others.

Conventional computational medicinal chemistry strategies—such as structure-based drug design, matched molecular pairs, positional SHAP, and Free–Wilson analysis—have been adapted to macrocyclic peptide programs, but they scale poorly in this regime.[10-12] Free–Wilson-style enumeration can systematically explore substitutions across multiple positions, yet it remains restricted to chemistries already tested within a given chemical series. In principle, one could enumerate all available monomers at each position, but for multi-residue macrocycles this rapidly produces libraries of billions to trillions of candidates, rendering exhaustive *in silico* scoring computationally impractical and often inefficient.

These limitations motivate the use of generative models to guide peptide design. Rather than exhaustively enumerating all combinations, generative approaches learn from existing data to propose new sequences or structures directly in promising regions of chemical space and can be coupled with property predictors and optimization algorithms to address multi-parameter objectives. In the following sections, we review existing generative models for peptide design, with a particular focus on methods aimed at property-driven lead optimization in macrocyclic peptide drug discovery.

## 1.2 **Overview of Generative Models in Peptide Design**

Generative models offer a data-driven framework for peptide design that addresses the combinatorial scale and multi-parameter nature of peptide optimization.[13] By learning statistical and chemical patterns from existing drug discovery peptide program data, these models can generate new sequences *de novo* or iteratively refine bona fide hits. This enables systematic exploration of large design (chemical and physico-chemical) spaces that are impractical to enumerate experimentally, while directly biasing generation toward desired physicochemical or functional property ranges.

Peptide generative models span multiple architectures, including variational autoencoders (VAEs),[14, 15] generative adversarial networks (GANs),[16] recurrent neural networks (RNNs), long short-term memory (LSTM) networks, diffusion models,[17] as well as transformer-based models.[18-23] Early generative models primarily focused on generating peptide sequences composed of canonical amino acids, typically represented in the FASTA format, such as Peptide-GPT[21] and PepGenWL.[20] More recent models extend beyond canonical amino acids by incorporating atom-level representations and NCAAs, enabled by improved datasets and model expressivity, such as PepINVENT,[18] PepTune,[17] PepThink-R1,[22] and HELM-GPT.[23]

In parallel, increasing interest in macrocyclic peptides has driven the development of structure-based generative approaches that operate directly in three dimensions. Methods such as AfCycDesign[24], BindCraft[25], and BoltzGen[26] integrate structure prediction[27, 28] and generative design[29] to produce peptide or protein binders. These approaches are currently limited to canonical amino acids and are more suited to early hit generation than to downstream lead optimization workflows.

The focus of this work is property-biased lead optimization using generative models after hit identification and validation. In this setting, localized and residue-specific modification is required. Most property-optimization frameworks follow a generator–predictor–optimizer paradigm, in which a generator proposes candidates, property predictors evaluate them, and optimization algorithms bias sampling toward favorable regions of sequence space. Among existing approaches, PepINVENT stands out as it enables controllable, monomer-level modification that aligns with medicinal chemistry design strategies. PepINVENT is a transformer-based generative framework for peptide design that integrates CHUCKLES,[30] a SMILES-like representation enabling atom-level control over both canonical and non-canonical amino acids, while still maintaining the monomer-level boundary. This allows medicinal chemists to keep specific residues that have steep structure-activity relationships (SAR) or are important for binding affinity, while focusing on optimizing the remaining positions for potency and/or property purposes. However, PepINVENT operates under several “static” assumptions: i) it requires chemists to pre-define which residues can be modified, a constraint that depends on prior structural or SAR knowledge and may overlook key positions; ii) its pretraining masks a fixed subset of residues, limiting exposure to diverse sequence contexts; and iii) the optimization phase treats the input peptide as static, preventing the model from learning dynamically from its own improved outputs. These limitations restrict the model’s generalization and flexibility, especially when optimal modification sites are unknown.

### 1.3 **Overview of Current Work**

In this work, we introduce new approaches to both the pretraining and optimization phases. During pretraining, we implement CHUCKLES shifting (**Figure 1-A**) and dynamic masking (**Figure 1-B**) to enable the model to learn richer and more comprehensive representations of peptides, ultimately improving its ability to generalize to unseen peptide sequences. We also introduce a context-free multi-armed bandit router (**Figure**

**1-C)** that adaptively identifies high-impact residues for editing without requiring chemists to pre-specify mutable positions. In the optimization phase, we propose a novel evolving algorithm that allows PepEVOLVE to learn from a broader range of inputs and contextual variations, rather than relying on a fixed input set as in PepINVENT (**Figure 1-D, 1-E**). Furthermore, we incorporate a group-relative advantage mechanism that enhances learning stability and encourages more effective exploration of the peptide design space, leading to more diverse and higher-quality peptide candidates.

## 2. Methods

### 2.1 **Motivations of Current Work**

**Pretraining Phase.** In PepINVENT, the pretraining process relies on a fixed masking dataset, meaning the masked positions in peptide sequences remain the same over training steps. This static setup limits the diversity of learning signals available to the model and increases the risk of overfitting. Since the model repeatedly encounters the same masked positions, it tends to memorize more than generalize, which weakens its performance on unseen peptide sequences. This issue is further compounded for cyclic peptides, which can be converted into multiple linear CHUCKLES representations depending on the chosen starting monomer; training with such a static and fixed sequence dataset can therefore further limit the model's ability to generalize.

**Optimization Phase.** During optimization with reinforcement learning (RL), PepINVENT keeps the input peptide sequence fixed across all iterations. This restriction prevents the encoder from learning effectively from variations in the input, and the model operates as if it has only a single input state, which limits its capacity to explore and adapt. We hypothesized that it is more effective to allow the peptide to change dynamically during optimization, letting it "evolve" by replacing the current input with improved peptides generated by the model itself. This approach enables the model to iteratively refine peptide sequences and explore the sequence space more effectively. However, the current implementation of PepINVENT limits such evolution. Even when the input sequence is updated with better residues during one of the reinforcement learning steps, the model still processes the sequence as if the evolved positions were masked, which prevents it from fully learning from the updated input.

To overcome these limitations, PepEVOLVE introduces a dynamic learning framework. During pretraining, the framework leverages dynamic masking and CHUCKLES shifting. In the optimization phase, an iterative optimization strategy enables peptides to adapt and evolve. Together these methods allow the model to learn, adapt, and generalize more effectively across a diverse peptide sequence space.

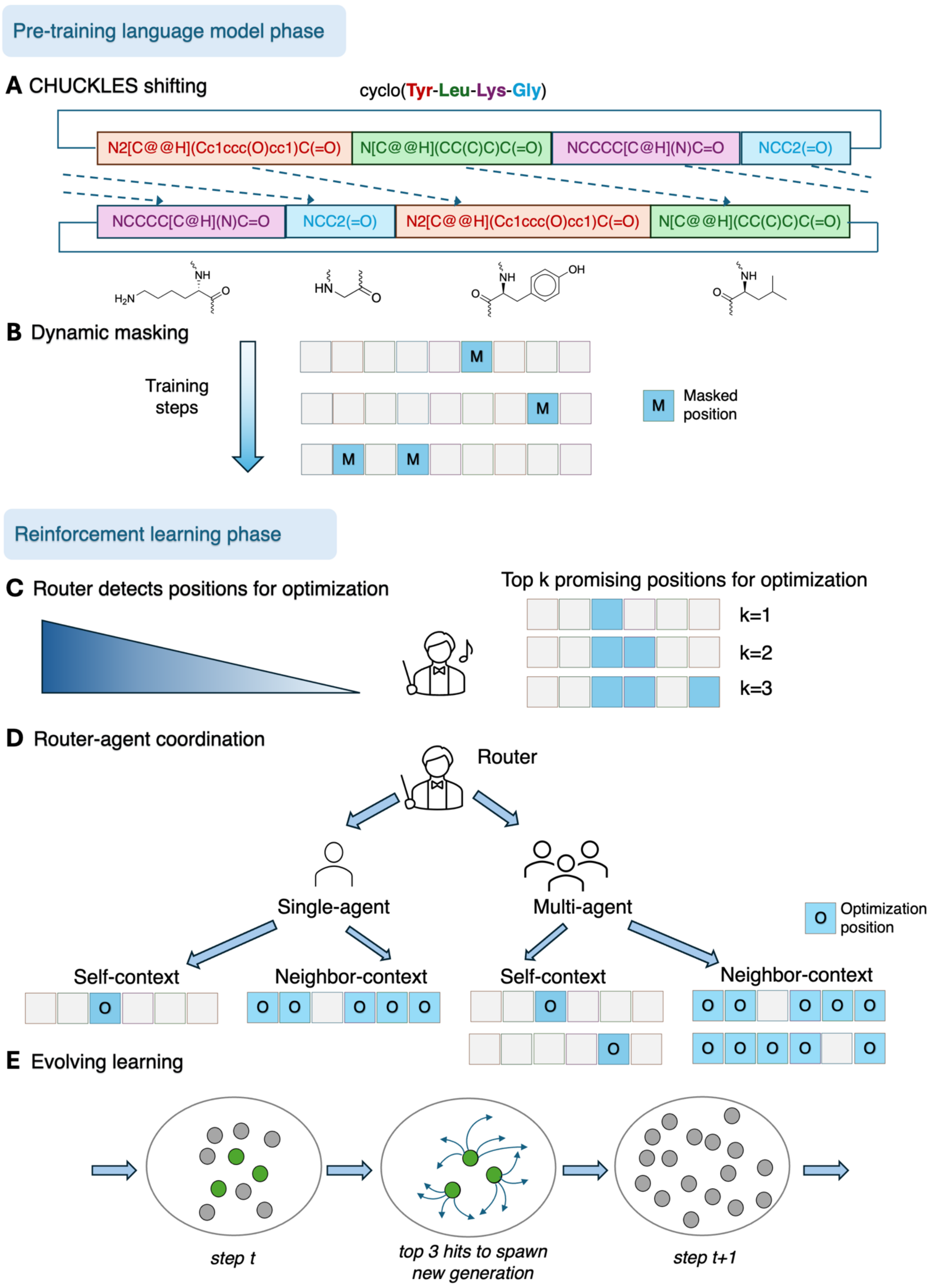

Pre-training language model phase
A CHUCKLES shifting
cyclo(Tyr-Leu-Lys-Gly)
N2[C@@H](Cc1ccc(O)cc1)C(=O)
N[C@@H](CC(C)C)C(=O)
NCCCC[C@H](N)C=O
NCC2(=O)
B Dynamic masking
Training steps
M
Masked position
Reinforcement learning phase
C Router detects positions for optimization
Top k promising positions for optimization
k=1
k=2
k=3
D Router-agent coordination
Router
Single-agent
Multi-agent
Self-context
Neighbor-context
O
Optimization position
E Evolving learning
step t
top 3 hits to spawn new generation
step t+1

**Figure 1. PepEVOLVE workflow.** PepEVOLVE combines language-model pre-training with a reinforcement-learning optimization loop to improve macrocyclic peptide generation. **(A)** *CHUCKLES shifting* augments the training set by shifting the linearized representation of a macrocycle, cyclo(Tyr–Leu–Lys–Gly), creating equivalent sequences that better reflect the cyclic topology. **(B)** *Dynamic masking* randomly masks different positions across the training steps, improving sample efficiency and reducing rote memorization. **(C)** In the RL phase, a *router* uses policy-gradient signals to rank residue/monomer positions and select the top *k* positions most likely to improve the objective. **(D)** The router then coordinates optimization using either a single agent or multiple agents, operating in *self-mask* (updating selected positions within a sequence) or *neighbor-mask* (coordinated updates across nearby/related positions). **(E)** *Evolving learning* iteratively retains the highest-scoring candidates at each generation step and uses them to seed the next round, progressively enriching the population toward improved monomer combinations.

## 2.2 **Model Architecture**

Our generative engine uses the same implementation and parameter configuration as the transformer architecture introduced in PepINVENT.[18] Briefly, the model is formulated as a conditional masked language model trained on peptide sequences in which selected positions are masked in the CHUCKLES representation, together with the corresponding ground-truth CHUCKLES strings for the missing amino acids. During generation, the model is conditioned on a masked peptide and predicts the missing tokens autoregressively, one token at a time, until the [END] token is produced or a maximum length of 500 tokens is reached. When multiple amino acid positions are masked, the model completes them sequentially: after predicting the first masked amino acid, the generated residue is incorporated into the context before prediction proceeds to the next masked position.

## 2.3 **Pretraining Peptide Language Model**

Our pretraining objective is to equip the generative model with residue level chemical understanding while remaining robust to representational shifts that arise from multiple linear representations of the same peptide. We build on PepINVENT's framework and retain its chemically grounded CHUCKLES encoding but modify the masking and sequence presentation, so the model learns context-aware peptide representations rather than memorizing static token patterns.

We trained and validated our models on the semi-synthetic peptide dataset introduced in PepINVENT.[18] The provided training and validation splits comprise ~900,000 and 50,000 peptides, respectively. Sequences are assembled from canonical and non-canonical alpha-amino acids, including building blocks from the virtual library developed by Amarasinghe *et al*. through reaction-based enumeration of eMolecules building blocks.[31] Additional diversity is introduced by stereochemical randomization, *N*-methylations, and both linear and cyclic designs.

Peptides are represented using CHUCKLES, a standardized SMILES-derived encoding that preserves atomic-level detail. Residues are concatenated in N-to-C order and separated by a “|” delimiter, yielding an atomistic string representation compatible with both canonical and non-canonical monomers. The dataset spans 6–18 residues and includes 40% linear and 60% macrocyclic peptides (i.e., 20% head-to-tail, 20% side-chain-to-tail, and 20% disulfide-bridged). Non-canonical residues account for approximately 30% of positions, and the original splits are stratified to preserve sequence-length and topology distributions across data splits (see Geylan *et al.* for full construction details).[18]

### 2.3.1 **Number of Masks Distribution**

During pretraining, PepINVENT uses a BERT-inspired text-infilling objective,[32] where 30% of amino acids in each peptide sequence are randomly masked. This objective trains the model to reconstruct the masked monomers, thereby learning peptide chemistry at the amino acid level. Because the shortest peptides in the pretraining data contained six monomers, a 30% masking ratio always resulted in more than one masked monomer per sequence.

In PepEVOLVE, we introduced a modified masking strategy to better align with the needs of our downstream optimization framework, where each agent must effectively handle single-residue optimization. The number of masked positions, $n_{\mathrm{mask}}$, was sampled from a triangular distribution[33] designed to bias the process toward single-site masking. We set the lower bound to be 1 (masking one amino acid) and the upper bound to be 0.4 $L$, where $L$ denotes the peptide length, which ensures the peak of the triangular distribution lands at one mask. (**Figure 2**).

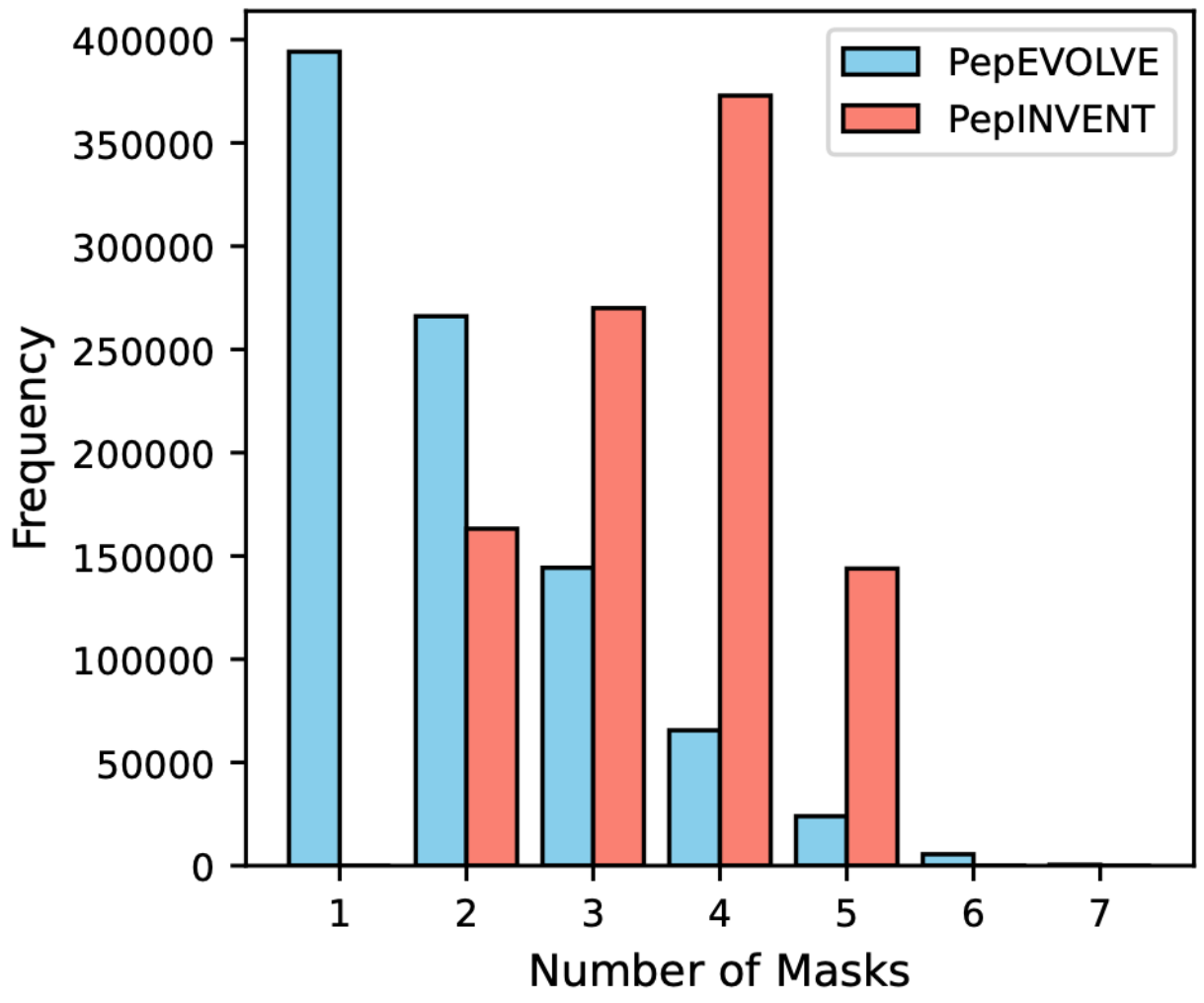

**Figure 2. Distribution of masked positions during pre-training.** Histogram of the number of masked residue positions (n_mask) sampled per training example for PepEVOLVE and PepINVENT (baseline masking) during pre-training. PepEVOLVE places greater probability mass on small n_mask values—especially single-position

masks—thereby emphasizing single-residue edits while still exposing the model to occasional multi-position masking for broader sequence remodeling.

### 2.3.2 **Pretraining with Dynamic Masking**

Building on the PepINVENT pretraining framework, PepEVOLVE introduces an enhanced text-infilling objective that integrates dynamic masking and CHUCKLES shifting to advance contextual learning and chemical robustness. In contrast to the fixed masking configuration employed by PepINVENT, PepEVOLVE implements dynamic resampling of masked positions at each epoch. This continual resampling exposes the model to a diverse array of context–mask combinations throughout training, thereby mitigating overfitting to specific residue locations and improving generalization of residue-level chemical relationships across variable local environments.

The CHUCKLES shifting mechanism further treats each peptide as a continuous chemical cycle rather than a fixed linear chain. Because cyclic peptides form closed loops without a defined start or end, their structural representation remains invariant under rotational shifts. Randomly shifting the CHUCKLES sequence thus enables the model to learn chemically equivalent representations of the same cyclic peptide, emphasizing intrinsic chemical connectivity rather than positional token order. For linear peptides, CHUCKLES shifting generates new samples, which serves as a data augmentation strategy, enhancing the model's ability to generalize across sequence variations.

To evaluate the performance of this pretraining strategy, the validation dataset was preprocessed in two ways:

1. a standard masked set, retaining the original sequences; and
2. a shifted masked set, where CHUCKLES shifting was applied prior to masking.

Models trained using static masking, as in PepINVENT, performed well on the standard set but exhibited a marked increase in validation loss on the shifted set, indicating sensitivity to positional ordering. Incorporating dynamic masking alone still does not alleviate this issue; the improvement in mean loss is small and still lies within the fluctuation of the static-masking baseline. In contrast, the combined application of dynamic masking and CHUCKLES shifting yields stable loss across both evaluation subsets, confirming that the model captures the underlying chemical principles governing peptide structure rather than memorizing sequence positions (**Figure 3**).

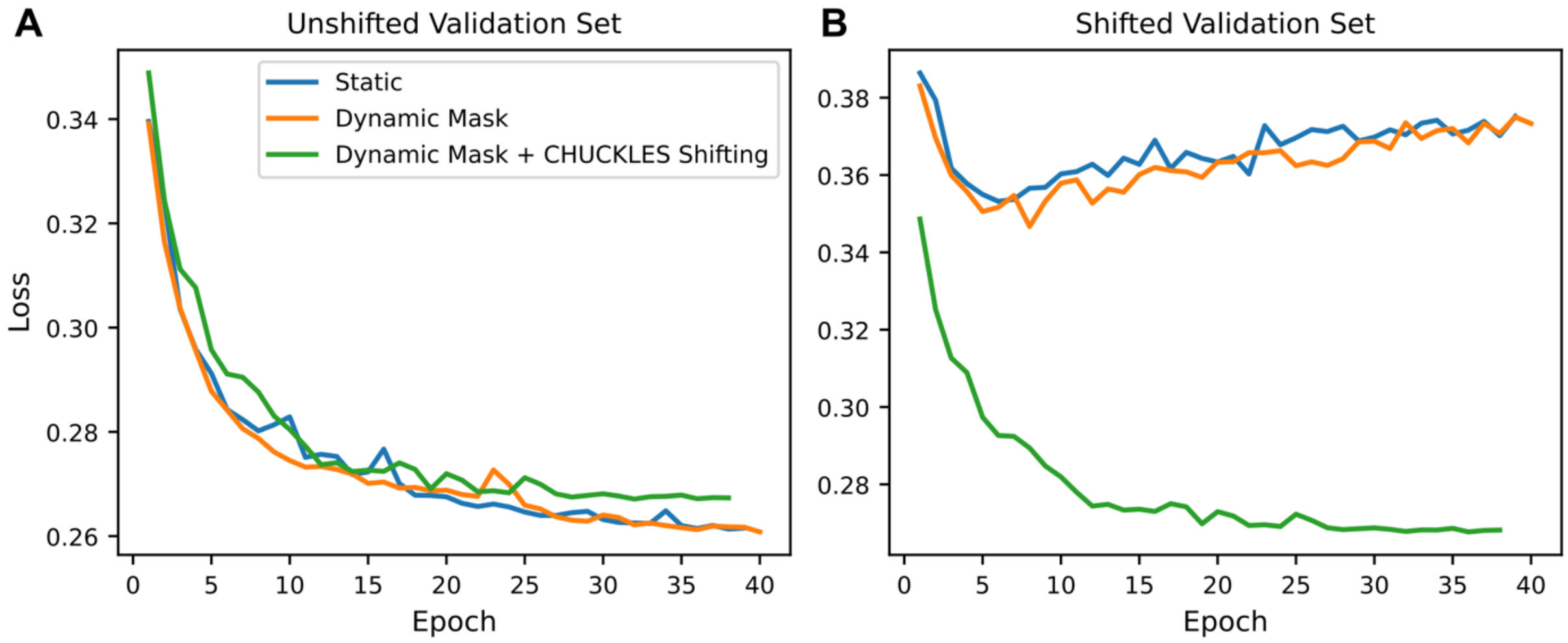

**Figure 3. Effect of dynamic masking and CHUCKLES shifting on pre-training generalization.** Validation loss across epochs for models trained with static masking, dynamic masking, or dynamic masking + CHUCKLES shifting. (A) Evaluation on the *unshifted* (standard linearization) validation set shows comparable convergence for all strategies. (B) Evaluation on the *CHUCKLES-shifted* validation set reveals a clear advantage for dynamic masking + CHUCKLES shifting, indicating improved invariance to cyclic-sequence re-indexing and better generalization to alternate macrocycle linearization.

## 2.4 **Router for Identifying Optimal Residues**

### 2.4.1 **Router Definition**

As discussed earlier, it is a challenging task for chemists to know a priori which residues are most promising for optimization. To automate this decision, PepEVOLVE introduces a router policy $\pi$, a high-level controller that governs the actions of the lower-level agent $f$. The router is formulated as a context-free multi-armed bandit that learns to prioritize residues solely from reward feedback of the peptides generated from the agent $f$, without conditioning on the peptide sequence.

Formally, a peptide ($p$) CHUCKLES with $L$ residues can be represented as

$$p = \text{"r}_1|\text{r}_2|...|\text{r}_L\text{"}$$

where each $\text{r}_i$ corresponds to the residue at position $i$. Following the PepINVENT framework, the agent $f$ requires explicit specification of the positions to be optimized, and this is achieved by introducing a special mask token, denoted “?”. For subsequent use, we define a masking operator $M(p, I)$, which replaces residues at the selected index set $I \subseteq [L]$ with the token “?”, thereby indicating the positions targeted for optimization.

$$M(p,I)[i] = \begin{cases} ?, & \text{if } i \in I, \\ \mathrm{r}_i, & \text{otherwise.} \end{cases}$$

The router policy is parameterized by $L$ variables, each indicating the probability of selecting a particular residue position within a categorical distribution, expressed as $\pi_\theta$ where $\theta \in \mathbb{R}^L$. During each routing step, the policy samples a batch $B$ of $K$ unique position indices to be optimized, these positions represent the subset of residues that is potentially optimal for modification. Each sampled subset is drawn from the categorical distribution governed by $\pi_\theta$ as follows:

$$[I]_{B\times K} = [i_{bk}]_{B\times K} \sim \pi_\theta, I \in [L]^{B\times K}$$

$$I_b = (i_{\{b1\}}, \dots, i_{\{bK\}}) \in [L]^K$$

For each selected subset $I_b$, a masked peptide

$$\tilde{p}_{I_b} = M(p, I_b)$$

is created and is then input into the agent $f$ to generate $G$ candidate peptides:

$$\{\hat{p}^g_{I_b}\}^G_{g=1} \sim f(\tilde{p}_{I_b}).$$

Each of these candidates receives an individual scalar reward $R^g_{I_b}$, and the mean reward across the $G$ candidates serves as the aggregate performance measure for the sampled subset:

$$\bar{R}_{I_b} = \frac{1}{G}\sum_{g=1}^{G} R^g_{I_b}.$$

This hierarchical design distributes control between two complementary levels. The router acts as a high-level controller, learning which residues are most valuable to modify. The optimizing agent operates at a lower level, performing the actual modifications at the selected positions and learning which specific changes yield the greatest improvement in reward. Together, this coordination enables PepEVOLVE to explore both where and how to modify peptides for optimal outcomes.

### 2.4.2 **Router Policy Update Mechanism**

The router policy is updated via a policy-gradient objective that leverages the aggregate rewards $R_{I_b}$ obtained for each sampled subset of positions. Specifically, the update follows the same REINFORCE-style[34] gradient estimator employed in PepINVENT, with the addition of an adaptive baseline to reduce variance. Given a batch of $B$ sampled subsets and their corresponding scores $R_{I_b}$, the advantage for each sample is computed as

$$A_{I_b} = R_{I_b} - \mathbb{E}[R_{I_b}]$$

where the expectation is approximated by a running exponential average, forming a moving baseline that stabilizes training. This baseline $\mathbb{E}[R_{I_b}]$ is updated after each step according to

$$\text{baseline} \leftarrow \lambda \, \text{baseline} + (1 - \lambda) \, \frac{1}{B} \sum_{b} R_{I_b}$$

where $\lambda \in [0,1)$ is the baseline smoothing coefficient.

The router assigns a log-probability to each sampled mask $I_b$ through its categorical distribution, $\log \pi_\theta(I_b)$. The policy gradient (PG) loss is then expressed as

$$\mathcal{L}_{\text{PG}} = -\frac{1}{B} \sum_{b} A_{I_b} \log \pi_\theta(I_b)$$

encouraging the router to increase the likelihood of residue subsets yielding higher-than-expected rewards.

To maintain exploration and prevent premature convergence, an entropy regularization term is additionally incorporated. The router's entropy $\mathcal{H}(\pi_\theta)$ quantifies its uncertainty over positions, promoting diversity in selected mutation sites. The final loss combines the policy gradient and entropy terms:

$$\mathcal{L}_{\text{router}} = \mathcal{L}_{\text{PG}} - \beta \, \mathcal{H}(\pi_\theta),$$

where the coefficient $\beta$ controls the degree of entropy regularization. To gradually shift from exploration to exploitation, $\beta$ is annealed linearly over training steps from an initial value $\beta_{\text{start}}$ to a final value $\beta_{\text{end}}$.

In summary, this update scheme allows the router to continually refine its probabilistic focus, allocating higher selection probabilities to residue positions that have historically yielded stronger downstream improvements. The combination of entropy annealing and baseline correction ensures a stable yet adaptive learning process that balances exploration of new mutation sites with exploitation of high-performing regions.

## 2.5 **Evolving with Varying Input Contexts**

Each input context is handled by a different agent. This configuration is advantageous when the output space is heterogeneous, where the relationships between contexts and their optimization dynamics differ significantly. By allowing each agent to specialize in its own context, they can better capture diverse optimization behaviors that would otherwise be averaged out in the single-agent setup. Self-mask agent is, in essence, an agent that specializes on a single position, analogous to how individual chemists focus on different residues/positions to optimize. The self-mask agent builds in-depth knowledge about how to optimize a specific position. The neighbor-mask agent is modifying other positions rather than the specified position (leave-one-out type of mask); this is analogous to adjusting multiple positions together to drive properties into an optimal property region.

To illustrate how these masking strategies and modes work in practice, a detailed example demonstrating both self-mask and neighbor-mask configurations for a toy peptide sequence is provided in the Supporting Information S1. For each input peptide $\tilde{p}^j$, $G$ candidate peptides $\hat{p}_g^j$ are generated. The candidates generated from each input peptide $\tilde{p}^j$ form a group, within which rewards are computed using a group-relative advantage mechanism.[35] To update the agent, the relative advantage of each candidate within its group is computed as:

$$A_g^j = \frac{R(\hat{p}_g^j) - \bar{R}^j}{\sigma_R^j + \varepsilon}$$

Here, $\bar{R}^j$ denotes the mean reward of all candidate peptides generated from the same seed $\tilde{p}^j$, representing the average performance within the group, and $\sigma_R^j$ represents the standard deviation of rewards within that group, which normalizes the advantage values to ensure scale invariance and stabilize learning across different reward magnitudes, and $\varepsilon$ is a small constant added for numerical stability. This enables learning to focus on the relative quality of candidates derived from the same seed peptide. Rather than comparing all candidates globally, each group formed from a common seed input serves as an independent reference frame. Within a group, candidates are evaluated based on how much better or worse they perform compared to their peers. Normalizing performance within each group provides a balanced and stable learning signal, ensuring that updates emphasize consistent improvement among all seeds.

To demonstrate the process, as illustrated in **Figure 4**, an example with the input peptide sequence "`A|B|C|D`", together with neighbor-mask context visibility, a set of optimal positions $O = \{0,2,3\}$, evolution parameters for top-K extraction (top number of seed $K = 2$ and number of candidates $G = 8$) is set for the optimization.

At the initial step (step 0), which is not shown in the **Figure 4**, $K = 2$ peptides are initialized from the model with the input "`?|B|?|?`" producing the seed inputs "`A|B|D|C`" and "`B|B|A|E`". Because there are $|O| = 3$ target positions to optimize, three distinct input contexts are defined, each processed sequentially by the agent. For every context, the agent handles $K = 2$ CHUCKLES, and for each CHUCKLES, the agent generates $G = 8$ outputs. These outputs are scored and normalized using the group relative advantage (GRA) method.[35] The resulting advantage values are then used to update the agent's weights, progressively refining its policy and performance.

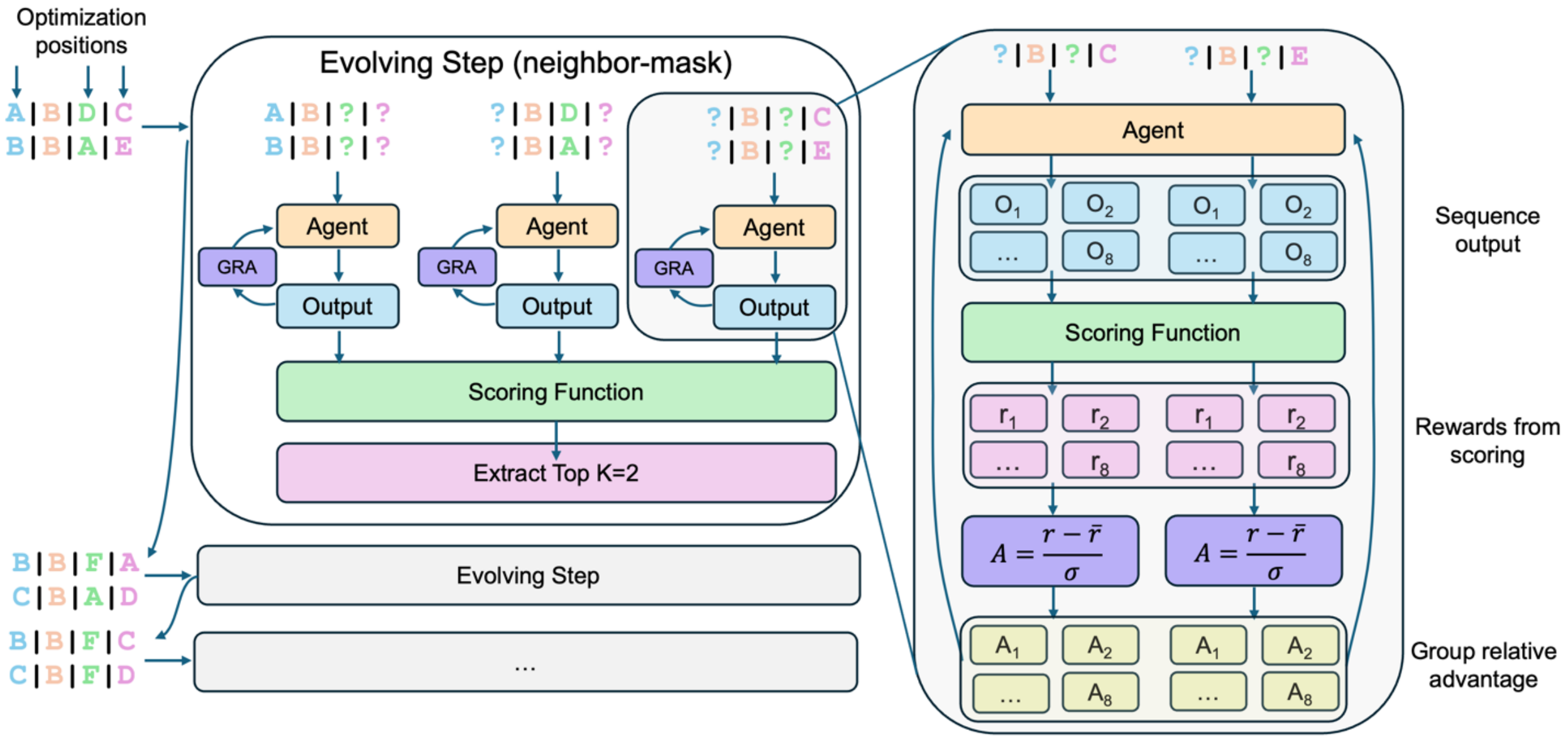

**Figure 4. Evolving-learning phase with group relative advantage (GRA).** The evolving phase iteratively refines peptides by reseeding generation from the best candidates in the current population. In the illustrated neighbor-masking example, the seed sequence "A|B|C|D" is optimized at positions $O = \{0, 2, 3\}$. At each evolving step, the top-K sequences from previous step (here $K = 2$) are selected as seeds. For each seed, $|O| = 3$ masked input contexts are extracted (one per optimized position and its local neighborhood). In every context, an agent samples G candidate completions per context (here $G = 8$). Candidates are evaluated by the scoring function to obtain rewards r, which are converted to group-relative advantages by within-group normalization. The resulting advantages are used to update the agent, and the next generation is formed by retaining the top-K scoring peptides, repeating the cycle.

After all peptides from the three contexts are generated, they are aggregated into a single pool and re-evaluated using similar scoring functions to get the raw scores, this time without normalization by group relative advantage (GRA). From this pool, the top $K = 2$ peptides are selected based on their computed scores to become new seed inputs for the next step of the evolving phase. This procedure continues iteratively, following the same algorithm throughout the entire evolving process.

By continuously exposing the model to a variety of inputs and contexts at every step, our evolving algorithm enables learning from a richer and more diverse input space. GRA plays a central role by guiding the model toward identifying which actions yield the greatest improvement. This iterative exposure and feedback allow the model to progressively refine its decision-making and improve performance throughout the evolving phase.

## 2.6 **Genetic Algorithm for Baseline Comparison**

To contextualize PepEVOLVE against a non-learning optimization strategy, we implemented a genetic algorithm (GA) baseline that operates directly on CHUCKLES-format peptides and reuses the identical scoring function as PepEVOLVE and PepINVENT. The GA optimizes the same four mutable positions used in the generative benchmarks. Its monomer pool is constructed from the pretraining data by counting interior monomer frequencies, and mutations sample from this pool proportionally to frequency.

Each generation applies tournament selection (k = 3), single-point crossover restricted to the mutable positions (probability 0.7), and per-position mutation (rate 0.15), with the top 10% of individuals carried over unchanged (elitism). Ring-closure digits are programmatically adjusted when terminal positions of macrocyclic peptides are mutated to preserve valid cyclic structures. We ran the GA with a population size of 128 for 1,000 generations, corresponding to 128,000 total scoring evaluations. This matches the evaluation budget of the generative methods: PepEVOLVE performs 250 evolving steps across 4 masked contexts (1,000 context iterations), each generating K = 16 seeds × G = 8 candidates, and PepINVENT performs 1,000 reinforcement-learning steps at a batch size of 128 — 128,000 scored peptides in every case.

## 3. Results

### 3.1 **Evaluating Router Effectiveness in Identifying Optimal Residues**

To evaluate the effectiveness of the router policy, we conducted experiments on three distinct test cases, each defined by an objective function with increasing complexity. The analysis focused on the learned probability distribution of the policy across all residue positions.

#### 3.1.1 **Hydrogen Bond Donor Reduction**

In this test case, we virtually created a linear peptide and arbitrarily placed two special amino acids at positions **13** and **15** (**Figure 5-A**). This represents a simple case and also shows that the model framework works for both linear and cyclic peptides.

**Position 13**: N[C@H](CNCCNCC(N)=O)C(=O), containing 4 hydrogen bond donors (HBD)

**Position 15**: N[C@@H](Cc1c[nH]c2ccccc12)C(=O)O, containing 3 HBD

To evaluate the routing policy’s performance, besides the two amino acids above, other amino acids in the peptide only had 1–2 hydrogen bond donors, and the objective function was defined to minimize the total number of HBD within the linear peptide. Therefore, the ground truth for the top optimal position is position 13, and the second optimal position is position 15. We examined two scenarios of identifying: (1) a single most optimal position, and (2) two most optimal positions.

In the single-position scenario, the routing policy successfully converged on position 13, with the probability distribution approaching 0.7 averaged over time, indicating a strong preference for that site as the optimal choice. In the two-position scenario, the policy effectively distributed probability between positions 13 and 15. Both were correctly identified as optimal sites, with the policy still displayed a slightly stronger preference for position 13, aligning with its higher HBD count.

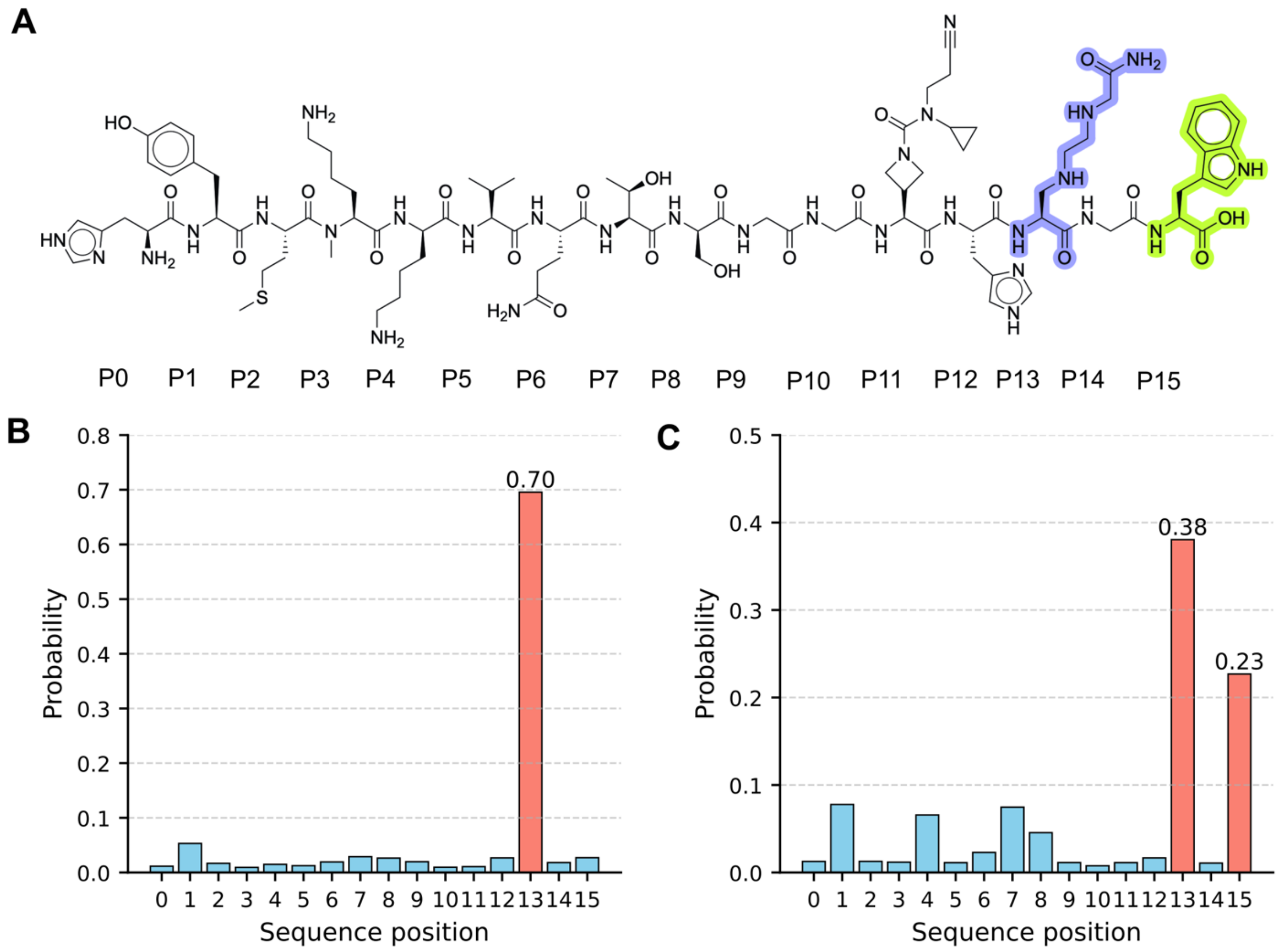

**Figure 5. Convergence of the router policy when optimizing hydrogen-bond donor (HBD) count.** (A) Reference peptide annotated by position (P0–P15); highlighted substituents indicate positions associated with reducing HBDs under the optimization objective. **(B, C)** Router selection probability (y-axis) as a function of sequence position (x-axis) when trained to minimize the total number of HBDs. **(B)** Top-1 routing (k = 1): probability mass concentrates on a single position, indicating convergence to one dominant edit site. **(C)** Top-2 routing (k = 2): probability mass concentrates on two positions, reflecting identification of a pair of complementary edit sites.

### 3.1.2 **cLogP Decrease**

In this test case, we used the head-to-tail macrocyclic peptide LUNA18[5] and focused on two specific amino acids that notably influence the overall cLogP value due to their aromatic side chains. We estimated cLogP using the Wildman–Crippen atom contribution method as implemented in RDKit.[36, 37]

**Position 4**: N(CC)[C@@H](Cc1ccc(C)cc1)C(=O), contributes moderately to an increase in cLogP

**Position 6**: N[C@@H](CCc1cc(F)c(C(F)(F)F)c(F)c1)C(=O), contributes most significantly to the cLogP increase.

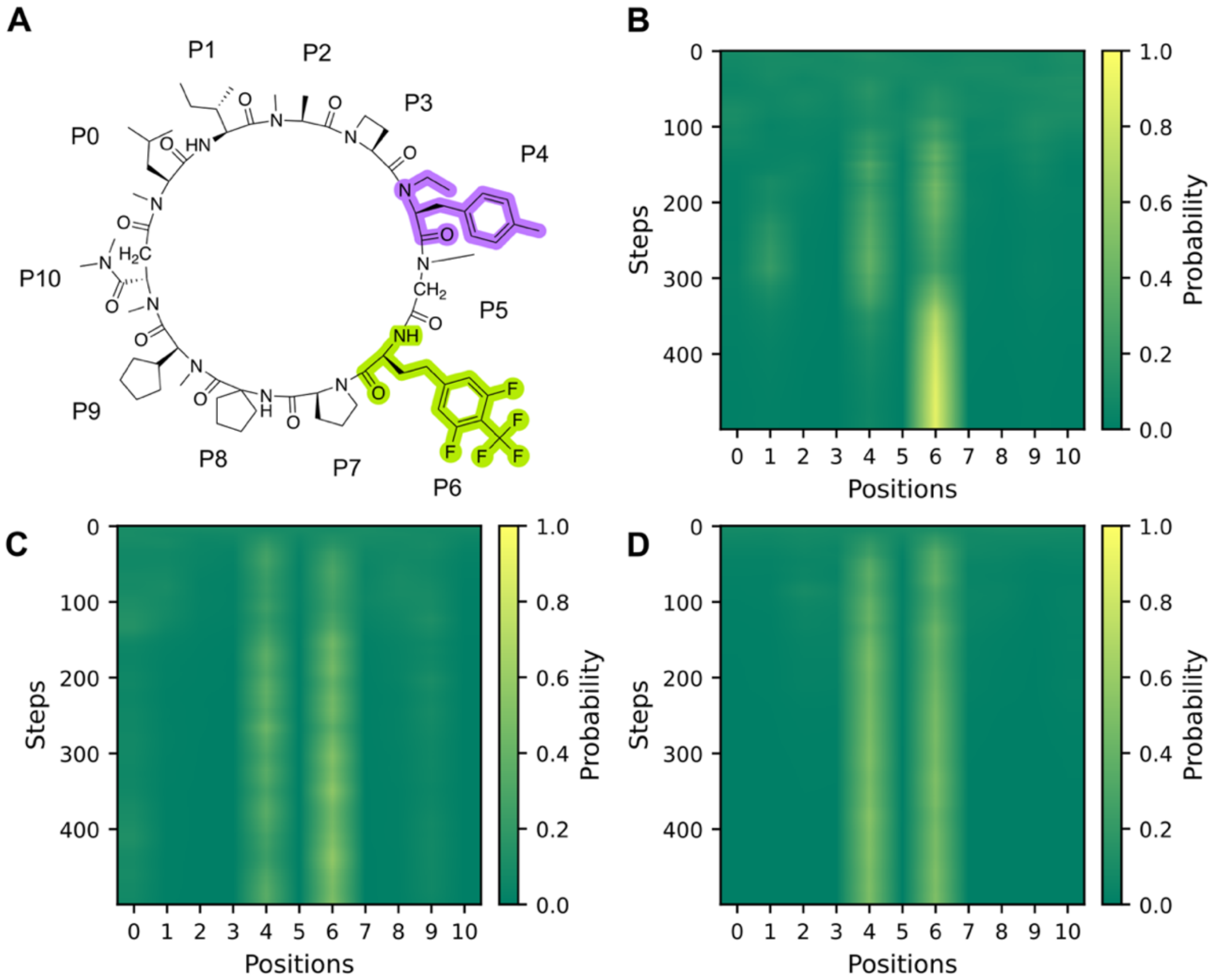

**Figure 6. Router policy convergence for cLogP optimization on the LUNA18 macrocycle. (A)** LUNA18 peptide scaffold with positions P0–P10; highlighted side chains indicate the primary sites the router converges on under cLogP-driven optimization. Heatmaps show router selection probability (color; 0–1) across routing steps (y-axis) and positions (x-axis). **(B)** Top-1 routing (k = 1): the router concentrates probability mass on a single position for decreasing cLogP. **(C)** Top-2 routing (k = 2): probability mass concentrates on two positions, indicating coordinated multi-site optimization for decreasing cLogP. **(D)** Adversarial control: starting from the lowest-cLogP peptide obtained in the LUNA18 setting, the objective is reversed (maximize cLogP); the router correspondingly shifts its selection distribution, demonstrating objective-dependent routing behavior.

Therefore, to assess the router policy's performance, the objective is defined as minimizing the overall cLogP of the peptide. The ground truth, therefore, corresponds to the two aromatic ring positions. As in the first test, we also evaluate the policy on two scenarios, identifying one and two optimal positions, over 500 routing steps. In the single-position experiment (**Figure 6-B**), the routing policy rapidly concentrated its probability mass on position 6, with values approaching 1.0 as training progressed. This indicates the model's strong confidence in selecting position 6 as the most influential contributor to the cLogP, consistent with the expected ground truth. When extended to identify two positions (**Figure 6-C**), the policy distributes its focus across positions 4 and 6. Both were accurately highlighted as key contributors, though position 6 maintains a marginally higher probability, aligning with its greater impact on the peptide's overall cLogP.

### 3.1.3 **Adversarial Task: cLogP Increase**

In this test case, we build upon the previous experiment, which identified the optimal positions for reducing the peptide's cLogP. As the router policy learns these optimal modification sites, the optimizing agent $f$ is also updated to generate better peptides with lower cLogP values. From the resulting set of generated peptides, the one with the lowest cLogP is selected, modifying positions 4 and 6. The corresponding starting peptide is provided in **Figure S2**. To evaluate the robustness of the router policy, the objective function is reversed, aiming at increasing the peptide's cLogP. The policy is then tasked with finding two optimal positions that would maximize the reward under this new objective. As illustrated in **Figure 6**-**D**, the router policy successfully converges on positions 4 and 6, matching the known ground truth (**Figure S2**).

## 3.2 Evaluating Optimization Capability: PepEVOLVE, PepINVENT, and GA

To evaluate PepEVOLVE against PepINVENT and GA, all the models were benchmarked on the therapeutically relevant Rev-binding peptide (RBP) used in PepINVENT's original study. Rev is essential in the HIV lifecycle and has been a target for antiviral discovery.[38] The RBP sequence *YPAASYR* was identified as a potential Rev inhibitor.[39] In the work by Wu *et al.,* two glycine residues were appended and a head-to-tail macrocyclization was performed to yield the macrocycle CHUCKLES.[40]

### 3.2.1 **Experiment Setup**

We evaluate four PepEVOLVE configurations: self-mask single-agent (SS), self-mask multi-agent (SM), neighbor-mask single-agent (NS), and neighbor-mask multi-agent (NM), as well as PepINVENT and GA under the same scoring function. Because the mutable sites are known, namely the alanines and the glycines (introduced solely to enable cyclization), routing was bypassed and direct evolution was performed. Each PepEVOLVE step processes one input context; for this task there are four target positions. To equalize total number of update steps across methods, we therefore run PepEVOLVE for 250 steps (4 positions $\times$ 250 = 1000 context iterations) and PepINVENT for 1000 steps. For the evolving phase, we use K=16 and G=8. To ensure robustness, we performed 3 random runs for each configuration.

Because the solubility code that was used in PepINVENT is not publicly available, we compare the three methods on the accessible scoring components: permeability, maximum ring size, lipophilicity, and custom SMARTS alerts (**Table S1**). All these scoring components are calculated consistent with PepINVENT's application. Permeability is predicted using the machine learning model provided in the PepINVENT repository. Maximum ring size is biased towards 14-30 atoms, favoring drug-like properties. Lipophilicity scores were biased to -5 ~ 0 for better solubility. The transformation and desirability functions are summarized in **Table S2**. To emphasize biological relevance, we aggregate scores using a geometric mean while giving higher weight to permeability (exponent = 3), as it is the primary optimization objective.

$$\text{Score} = (S_{\text{permeability}}^{3} \times S_{\text{ring}} \times S_{\text{lipophilicity}} \times S_{\text{SMARTS}})^{1/6}. \quad (Eq\ 1)$$

This makes the total scale interpretable while preserving multiplicative penalties when any component scores poorly.

### 3.2.2 **Score Distribution of Generated Peptides and Optimization Dynamics**

From the output of the 3 random runs combined, we summarize the score distribution of three methods over unique peptides (**Figure 7-A**) and the mean score of generated peptides along the optimization steps (**Figure 7-B**). For PepEVOLVE, across all four configurations, it clearly outperforms both PepINVENT and GA baseline. PepINVENT's generated peptide scores distribution concentrates around ≈ 0.6 with little mass beyond 0.8, and its best generated sequence has a score of ≈ 0.89. In contrast, PepEVOLVE shifts the bulk of the distribution upward: means are ≈ 0.8 with heavy tails extending to ≥ 0.9–0.95. Optimization dynamics mirror this: PepEVOLVE reaches high-scoring regions in fewer steps and maintains higher means throughout.

Among PepEVOLVE variants, SS exhibits the fastest early climb (rapidly entering the ≥ 0.9 region), SM attains the highest terminal scores, NS tracks closely behind with a stable ascent, and NM lags the others but still surpasses PepINVENT by a wide margin. Additionally, SS also exhibits a score drop around 800–900 steps. This is because the score is weighted by uniqueness, where the score from unique molecules is divided by the total number of generated samples, so a decreasing number of unique molecules is reflected in the score.

Because the GA population is initialized from the query peptide with each generated by randomly mutating one or more of the mutable positions, the GA begins optimization from a lower-scoring population than the generative models, which sample from a trained prior. This is reflected in the long tail of low-scoring samples in **Figure 7-A**. In terms of optimization dynamics, GA and the best-performing PepEVOLVE configurations both raise their scores rapidly in the early steps, but the GA quickly becomes trapped in a local optimum and plateaus well below the scores reached by PepEVOLVE (**Figure 7-B**).

Although GA was able to generate high-scoring peptides (see the trajectory of highest scores along optimization steps at **Figure S7**), it struggles to generate a diverse set of high-scoring peptides as PepEVOLVE does.

To understand the optimization behavior of PepEVOLVE configurations, we examine the number of unique SMILES (**Figure 7-C**), the average chemical diversity (**Figure 7-D**), and chemical space trajectories **(Figure 8-A, Figure S3)** throughout RL steps. In terms of uniqueness and diversity, neighbor-mask configurations sustain high molecular uniqueness and diversity throughout RL steps, whereas SS undergoes an early mode collapse beginning around 300 steps, with SM starting to show decline around 700 steps (Supporting Information S2). Despite this decline of uniqueness, SS and SM continue to yield high-scoring peptides right until around 900 steps. This dynamic can be explained by examining how each configuration traverses the chemical space

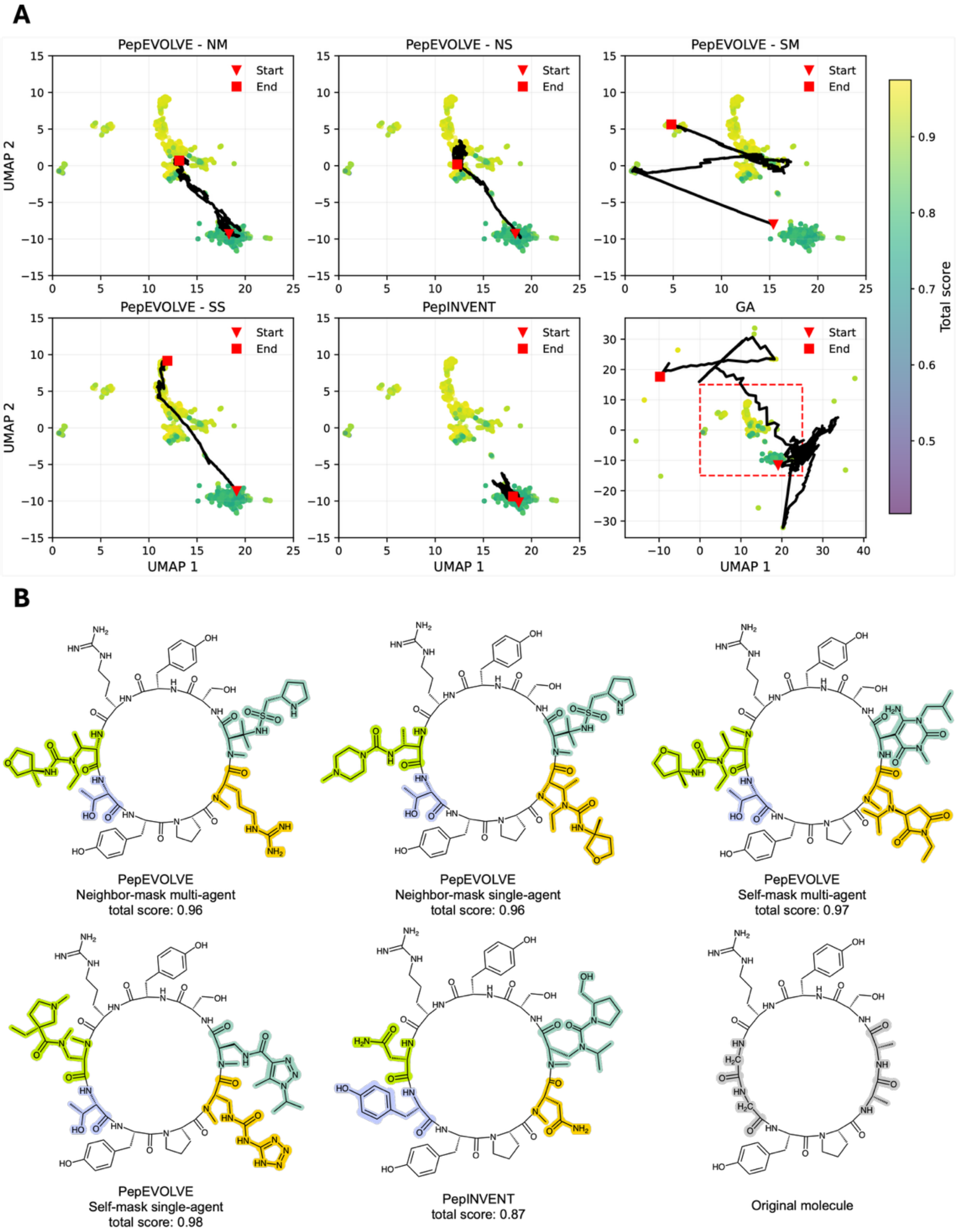

in

**Figure 8. Optimization trajectories in chemical space and representative optimized macrocycles.** (A) UMAP42 embedding of generated molecules using count-based Morgan fingerprints,[43] colored by total multi-objective score (color bar). For each method (PepEVOLVE NM/NS/SM/SS, PepINVENT, and GA), the black polyline traces the step-by-step location of the best-scoring molecule (elite) across optimization steps, illustrating how optimization progresses through chemical space. Red markers denote the initial molecule (triangle) and the final molecule (square). Non-GA methods are zoomed in; the zoomed-in region is indicated by a dashed box in the GA panel. (B) Structures of the top-scoring macrocycles produced by each method for the Rev-binding optimization task, shown alongside the original starting molecule; captions report the corresponding total scores. Relative to PepINVENT, PepEVOLVE variants identify higher-scoring solutions and explore more structurally elaborated chemotypes. The top-scoring molecule from GA can be found in the Figure S4.

PepINVENT remains largely confined to its initial chemical space; among the four PepEVOLVE configurations, NM and NS move toward the same high-scoring region of chemical space, while SM and SS find distinct regions of high-scoring yet cliff-like chemical space. This signals that the self-mask configurations can trade off uniqueness for aggressive exploration of distant high-score cliffs, whereas neighbor-mask configurations strike a more stable balance between exploiting and exploring the chemical space.  For the GA baseline, the number of unique SMILES at each step stays essentially constant, as it mutates and crosses over a fixed, pre-specified population size. Its cumulative diversity, however, remains low, since it cannot propose chemical matter beyond its existing monomer pool. This limits its escape from local optima, consistent with its early plateau in **Figure 7-B**.

### 3.2.3 Assessing Top-Quality Generated Peptides

We visualize the peptides to inspect the motifs generated by the models, including the original peptide, PepINVENT top total score peptide, and the top representative peptides from the four modes of PepEVOLVE (**Figure 8-B**). Using the same number of total steps, PepINVENT generates relatively simple monomers for the four positions, while PepEVOLVE generates more elaborate monomers at these positions. All the models commonly use amide and urea motifs to lower cLogP, and use *N*-methylation to improve permeability. Some PepEVOLVE-generated peptides also explore sulfonamide motifs to optimize towards desired cLogP. Overall, the dynamic optimization in PepEVOLVE allows the model to have more capacity to generate novel and bespoke chemical matter.

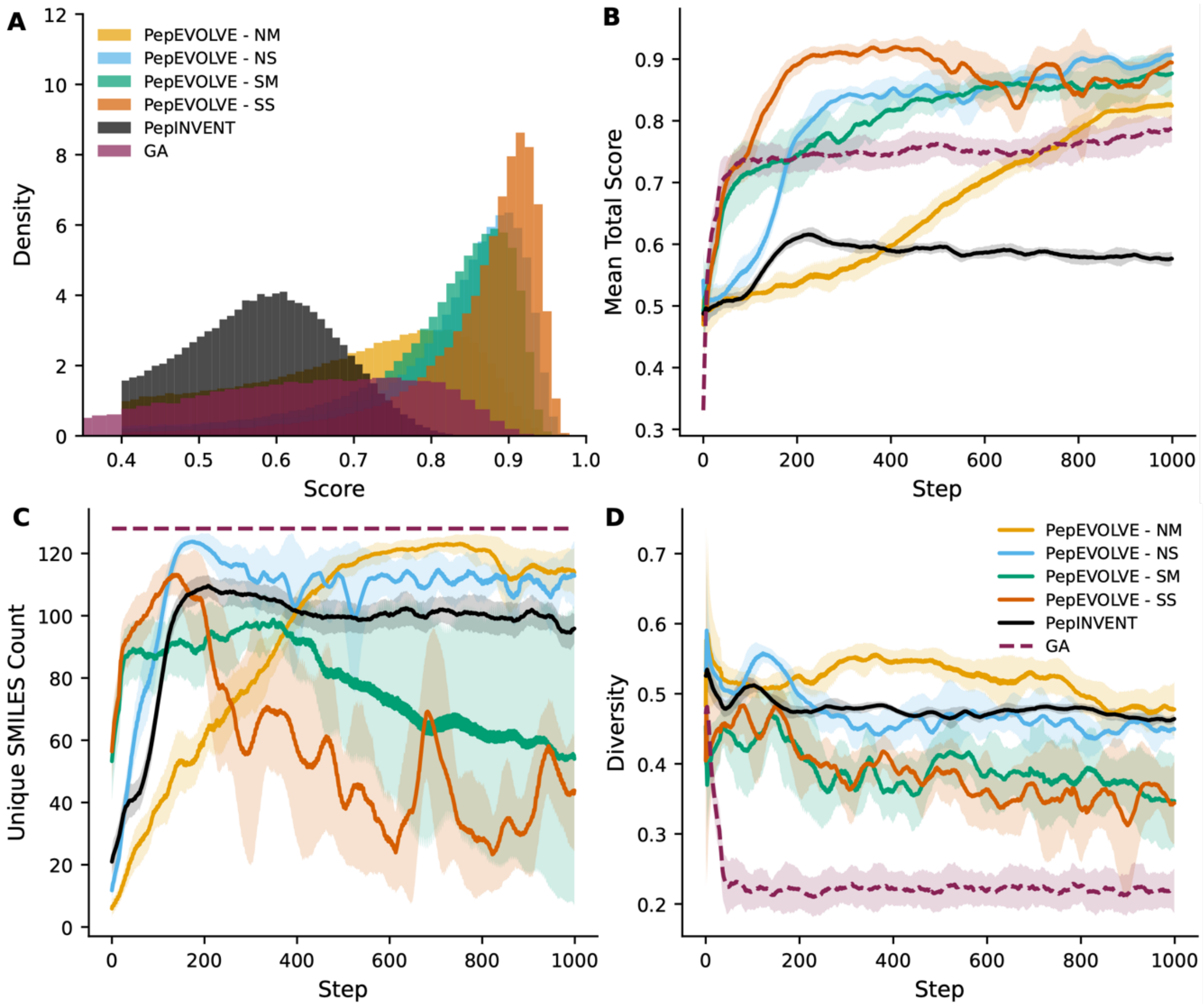

**Figure 7. PepEVOLVE, PepINVENT, and GA on Rev-binding macrocycle optimization under a permeability-aware multi-objective.** Performance of GA, PepINVENT and four PepEVOLVE variants—neighbor-mask multi-agent (NM), neighbor-mask single-agent (NS), self-mask multi-agent (SM), and self-mask single-agent (SS)—evaluated with the permeability-aware scoring function (Eq 1). (A) Distribution of generated peptide scores (higher is better). GA includes a long tail of low score samples and is truncated from the figure for clarity. (B) Average generated peptide scores over the optimization steps. (C) Average number of unique SMILES generated at each RL step. (D) Average chemical diversity of generated molecules at each RL step. Chemical diversity is calculated by 1 – mean Tanimoto similarity[41] over the generated set (higher values indicate greater diversity). Shaded regions indicate variability across runs (s.d.). A rolling average (window=30) is applied to help visualization.

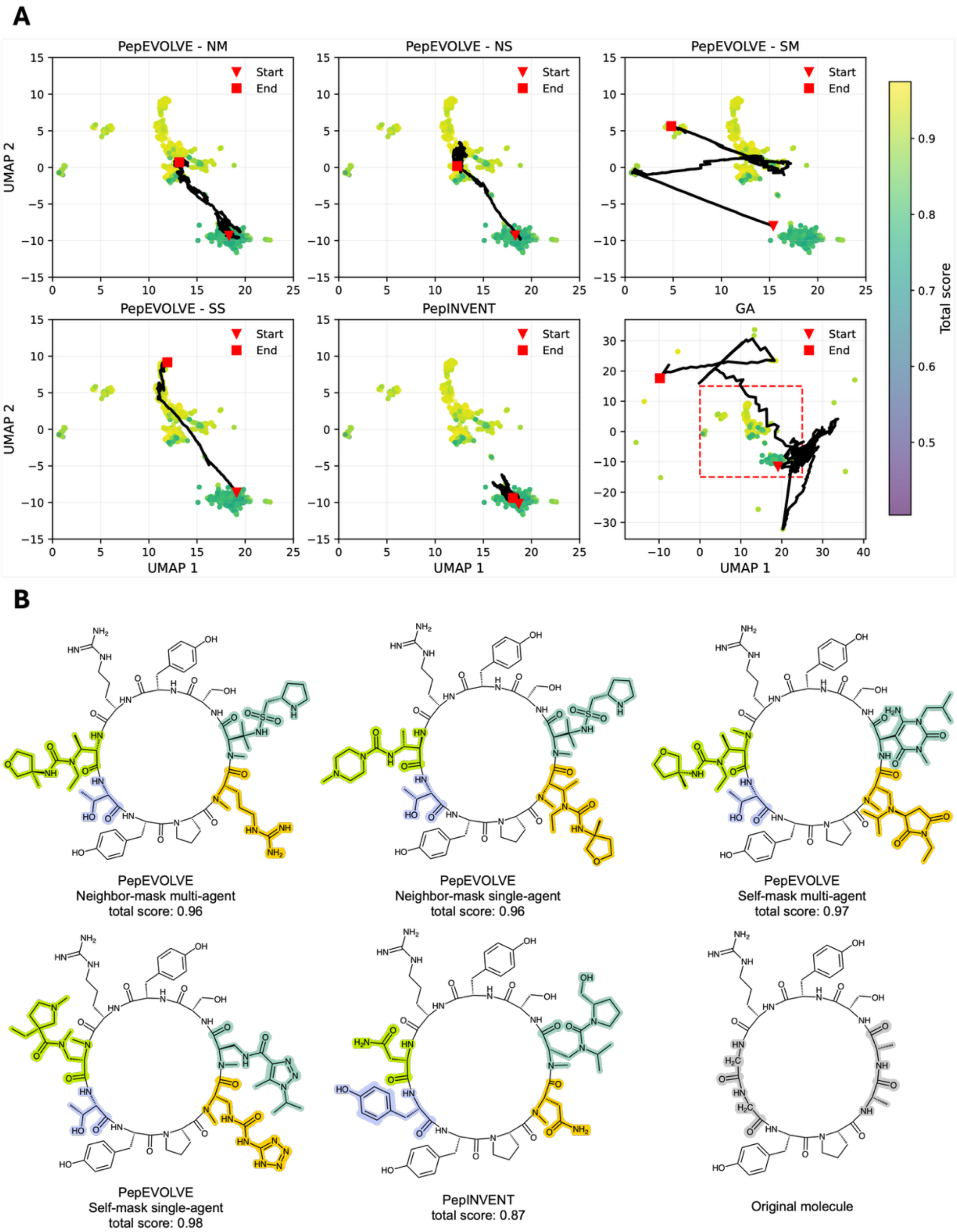

**Figure 8. Optimization trajectories in chemical space and representative optimized macrocycles.** (A) UMAP[42] embedding of generated molecules using count-based Morgan fingerprints,[43] colored by total multi-objective score (color bar). For each method (PepEVOLVE NM/NS/SM/SS, PepINVENT, and GA), the black polyline traces the step-by-step location of the best-scoring molecule (elite) across optimization steps, illustrating how optimization progresses through chemical space. Red markers denote the initial molecule (triangle) and the final molecule (square). Non-GA methods are zoomed in; the zoomed-in region is indicated by a dashed box in the GA panel. (B) Structures of the top-scoring macrocycles produced by each method for the Rev-binding optimization task, shown alongside the original starting molecule; captions report the corresponding total scores. Relative to PepINVENT, PepEVOLVE variants identify higher-scoring solutions and explore more structurally elaborated chemotypes. The top-scoring molecule from GA can be found in the Figure S4.

### 3.2.4 Yield-Quality Trade-Off

To validate the efficiency of generating high-quality peptides, we partitioned the unique peptides combined from 3 runs into different score bins, as shown in **Figure 9**. PepINVENT generates many unique sequences, but they are dominated by ≤0.8 scores. The GA baseline generates by far the fewest unique peptides, and a large fraction of these fall in the lowest score bins (≤0.4), with virtually no candidates above 0.8. In contrast, all PepEVOLVE variants shift their mass into the 0.80–0.94 bin, with significant counts above 0.9 bins, a regime not reached by PepINVENT. Among the variants, NS achieves the largest overall yield, exceeding PepINVENT, while keeping a substantial fraction in the 0.90–0.94 bin; SS produces fewer total sequences than NS but the highest proportion in the top bins; SM balances yield and quality; and NM, while generating many sequences, still rebalances markedly toward higher scores relative to PepINVENT. Overall, PepEVOLVE models generate a substantially higher fraction of peptides with scores between 0.8 and 1.0. Taken together, PepEVOLVE improves both sample efficiency (fewer steps/oracle calls to high scores) and the quality distribution without sacrificing diversity of unique outputs.

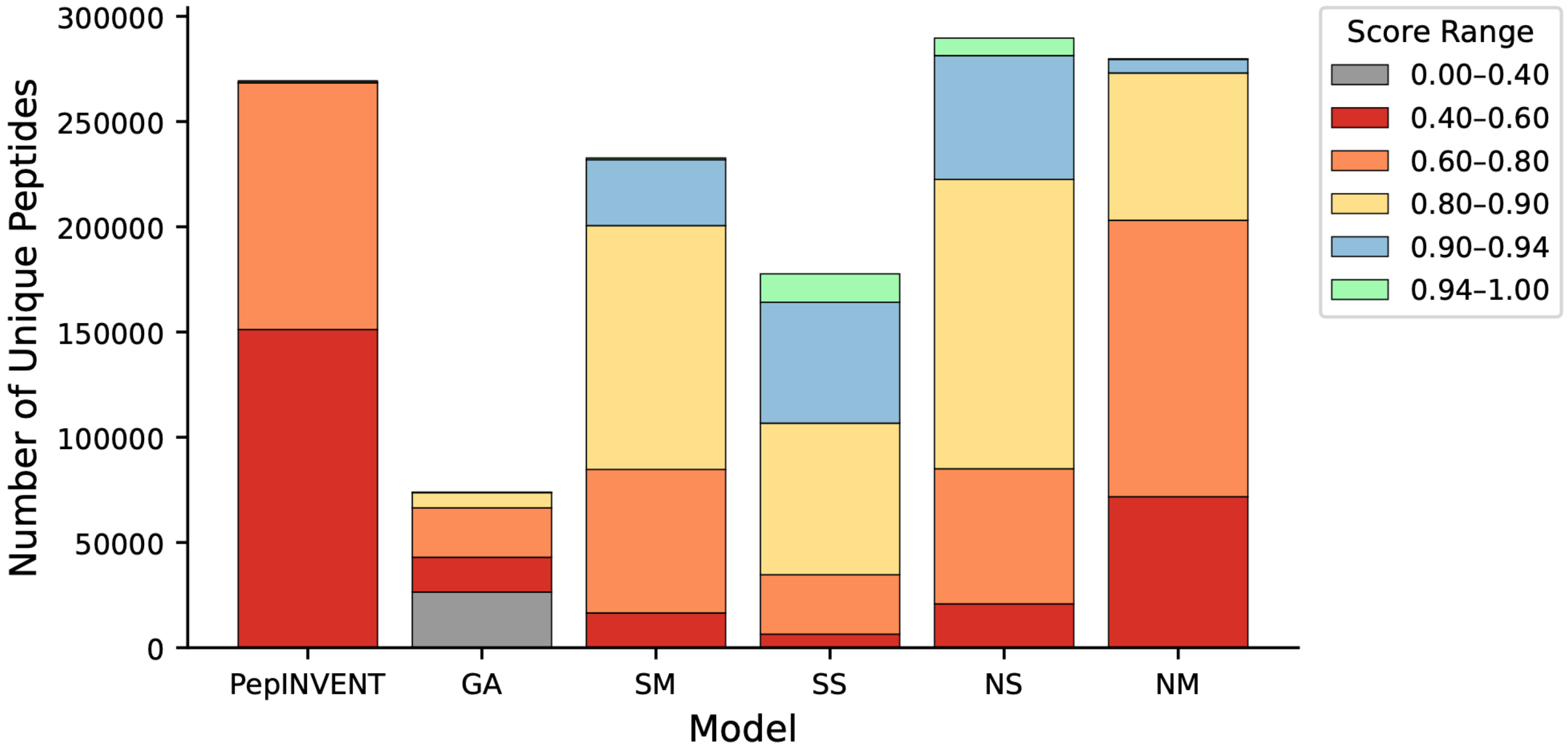

**Figure 9. Score-stratified yield of unique peptides across models.** Stacked bar chart showing the number of unique peptides generated by GA, PepINVENT and four PepEVOLVE configurations (SM, SS, NS, NM), grouped into total-score bins (0.00–0.40,

0.40–0.60, 0.60–0.80, 0.80–0.90, 0.90–0.94, 0.94–1.00). Relative to PepINVENT and GA, PepEVOLVE shifts the distribution toward higher-scoring peptides, producing a larger share of candidates in the 0.80–1.00 range.

### 3.2.5 Ablation Study to Quantify Contributions of PepEVOLVE Core Components

We conducted an ablation study to quantify the individual and combined contributions of PepEVOLVE's core components relative to the PepINVENT baseline (Supporting Information S3). Consistent with the learning curves and score distributions, both the baseline PepINVENT model and the CHUCKLES shift–only configuration achieved the lowest mean total scores and generated peptides predominantly concentrated in mid-scoring regions (**Figure S5**). Introducing the Top-K extraction mechanism markedly improved performance, broadening the score distribution and increasing the fraction of higher-scoring sequences. The largest gain was observed when Top-K extraction was combined with Group-Relative Advantage (GRA), which rapidly increased mean total scores and shifted the bulk of generated peptides into a sharply peaked, high-scoring regime (**Figure S6**). Across all agent configurations, the full PepEVOLVE framework (CHUCKLES Shift + Top-K + GRA) consistently achieved the highest and most stable performance (**Table S3, S4**), indicating that these components act synergistically to reliably generate a large volume of high-quality peptide candidates.

### 3.2.6 Summary

Across configurations, PepEVOLVE consistently outperforms PepINVENT in the following aspects: (i) higher mean and best-case scores, (ii) faster convergence to the high-scoring region, and (iii) a favorable yield–quality profile with many more peptides above 0.8 (and sizeable counts above 0.9). These gains hold under a permeability-aware objective and are obtained with fewer effective steps, demonstrating both efficiency and robustness.

Given there are different modes in PepEVOLVE, for practical applications, we recommend choosing the configuration to match the goal and compute budget:

- **Exploration or trend searching.** Start with neighbor-mask, single-agent (NS). It is the most stable across steps, since the single agent oversees all the positions. This yields the largest set of unique peptides and provides consistently strong scores while covering chemical space.
- **Top-hit search.** If the priority is to surface the highest-scoring candidates, start with self-mask single-agent (SS). It reaches the 0.9–1.0 band fastest and produces the highest fraction of elite peptides, at the cost of fewer uniques and a higher risk of late-stage mode collapse.
- **When to use multi-agent.** If scores widely fluctuate, different masking contexts might have significant optimization dynamics, switch to the corresponding multi-agent variants (SM or NM). These capture context-specific dynamics that a single agent may average out.

## 4. Discussion

This work introduces PepEVOLVE, a position-aware, dynamic optimization framework that augments a transformer generator with (i) dynamic pretraining (dynamic masking and CHUCKLES shifting), (ii) a router policy to automatically identify high-value residue positions with high probability of success, and (iii) an evolving optimization loop with group relative advantage (**Table 1**). Collectively, these innovations yield faster convergence and bespoke monomer designs than PepINVENT on a therapeutically motivated benchmark, with mean scores around 0.8 and best designs approaching 0.97 under a permeability-weighted geometric scoring function, while PepINVENT's best design scores reach near ~0.8 at best (**Figure 7-A**). Additionally, the comparison with a genetic algorithm baseline, run at the same scoring-evaluation budget as PepEVOLVE, highlights the advantage of generative modeling over enumeration-based methods that draw from a fixed monomer pool: the GA plateaus below 0.8, whereas PepEVOLVE reaches best scores near 0.97. In other words, generative method enables access to more privileged, higher-scoring chemical space. PepEVOLVE provides a practical path for lead optimization when medicinal chemists either do not know a priori which positions to change or wish to explore broader local neighborhoods efficiently.

Generative peptide design has progressed from residue-level VAEs, GANs, and RNNs to transformer and LLM-based models with NCAA-capable encodings.[14, 16, 18-23] PepINVENT established property-biased reinforcement learning over CHUCKLES, but requires users to pre-specify mutable positions, which risks missing more influential sites. Multi-agent frameworks (e.g., MultiMol,[44] MT-Mol[45]) highlight the value of role specialization in molecular design. PepEVOLVE advances this line by (a) removing static assumptions in pretraining via dynamic masking and rotationally invariant CHUCKLES shifts that stabilize validation loss across shifted/unshifted inputs, and (b) introducing a context-free bandit router that learns where to edit from reward alone, no sequence conditioning or hand-picked sites required.

**Table 1**. Summary of the advantages of PepEVOLVE.

| Component | PepINVENT | PepEVOLVE | Benefit |
|---|---|---|---|
| Masking | Static | Dynamic triangular | More context variety |
| CHUCKLES | Static | Shifting | Invariance |
| Editing positions | Manual | Router bandit | Automatic site discovery |
| Optimization | Static RL | Evolving multi-step | Dynamic refinement |
| Advantage | Vanilla | GRA | Stability |

**Limitations.** Our comparative metric removes solubility due to unavailable code and instead uses a permeability-weighted geometric mean (a weight of 3 for permeability and a weight of 1 to all other objectives) over ring size, lipophilicity, and SMARTS alerts; conclusions therefore reflect the chosen surrogate objective rather than full developability.

Benchmarking centers on one public case (Rev-binding macrocycle with predefined mutable A/G positions), with 250 PepEVOLVE steps versus 1000 for PepINVENT to equalize context passes; broader targets and alternative step budgets should be tested. The router is context-free (sequence-agnostic), which may miss sequence-dependent or 3D-structural contingencies. Finally, group-relative advantage, while stabilizing across local groups, can reduce diversity via group-wise normalization pressure.

## 5. Conclusion

PepEVOLVE addresses a key challenge in macrocyclic peptide design: identifying where to modify a sequence and guiding those changes so multiple properties improve simultaneously. By pairing dynamic pretraining (mask resampling and CHUCKLES shifting) with a router that learns high-value positions and an evolving, group-relative advantage optimizer, the framework shifts from static, position-guessed edits to data-driven, position-aware optimization. Across synthetic objectives and the RBP benchmark, PepEVOLVE consistently produced higher-scoring peptides and reached strong solutions in fewer steps than PepINVENT, while the router reliably found the positions most aligned with the objective. This approach closes the loop between medicinal chemist intuition and algorithmic search, it reduces exploration in an enormous design space and makes lead optimization more rapid, systematic, transparent, and reproducible, helping advance the state of computational peptide design.

## Supporting Information

The Supporting Information includes the following content.

**S1.** Example of Masking Strategies.

**S2.** Score Analysis and Model Collapse Behavior Observed.

**S3.** Ablation Study of PepEVOLVE Components.

**Figure S1.** Generated peptide scores over the steps when the mean score is calculated using all generated peptides rather than only unique ones.

**Figure S2.** Starting peptide used for router policy convergence in the adversarial test.

**Figure S3.** Chemical space analysis for PepEVOLVE vs. PepINVENT on Rev-binding macrocycle optimization.

**Figure S4.** Top-scoring molecule from GA run for RBP case.

**Figure S5.** Learning curves demonstrating the optimization performance of various ablation configurations.

**Figure S6.** Score density distributions of generated peptides across different ablation configurations.

**Figure S7.** Highest total score of generated peptides from each optimization step on Rev-binding macrocycle optimization under a permeability-aware multi-objective.

**Table S1.** SMARTS alert patterns for RBP optimization.

**Table S2.** Transformation and desirability functions for RBP optimization.

**Table S3.** Highest score of the peptide achieved across the PepINVENT baseline and various ablation configurations.

**Table S4.** Average scores and standard deviations of the top 10 generated peptides across the PepINVENT baseline and various ablation configurations.

## Acknowledgement

We are grateful to James P. Jewell, Edward Miller, and Ryan Chau for their chemists' intuition and insights. We thank Ruheng Wang, Ruibo Zhang, Yixiang Mao, Alec Glisman, Haote Li, Kenneth López Pérez, and Song Yin for valuable discussions on peptide prediction and generation methodologies. We thank Jennifer M. Johnston, Deeptak Verma, and Viktor Hornak for helpful discussions and departmental support.

## Data Availability Statement

The PepEVOLVE project is available at https://github.com/MSDLLCpapers/PepEvolve/. The repository provides installation instructions, input configuration files for the manuscript experiments, and scripts to perform pre-training and generation. Due to large file sizes, the pretrained model weights (https://huggingface.co/Merck/PepEVOLVE) and sample output for generated peptides with different PepEVOLVE and PepINVENT configurations (https://huggingface.co/datasets/Merck/PepEVOLVE-RBP-Benchmark) are available on HuggingFace.

## Conflict of Interest

T.N. was employed as an intern at Merck Sharp & Dohme LLC, a subsidiary of Merck & Co., Inc., Rahway, NJ, USA, during this work. P.Z.Z., N.B., C.F., L.Z., S.S., H.-W.P., and S.F. were full-time employees of Merck Sharp & Dohme LLC, a subsidiary of Merck & Co., Inc., Rahway, NJ, USA, at the time of the research and may hold stock and/or stock options in Merck & Co., Inc., Rahway, NJ, USA.

## TOC Figure

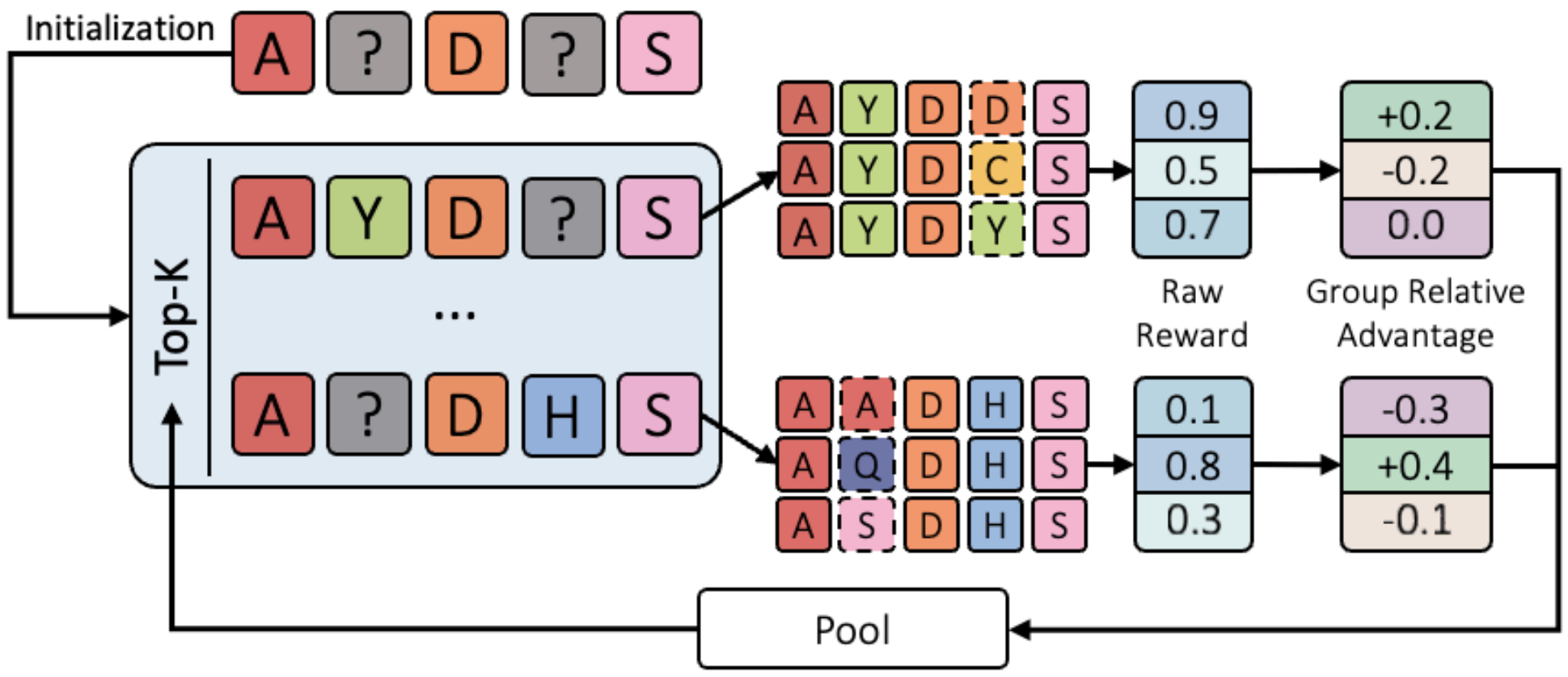
Initialization
A ? D ? S
Top-K
A Y D ? S
...
A ? D H S
A Y D D S
A Y D C S
A Y D Y S
0.9
0.5
0.7
Raw Reward
+0.2
-0.2
0.0
Group Relative Advantage
A A D H S
A Q D H S
A S D H S
0.1
0.8
0.3
-0.3
+0.4
-0.1
Pool

## References

(1) Moiola, M.; Memeo, M. G.; Quadrelli, P. Stapled Peptides—A Useful Improvement for Peptide-Based Drugs. In *Molecules*, 2019; Vol. 24, p 3654.
(2) Cardote, T. A.; Ciulli, A. Cyclic and Macrocyclic Peptides as Chemical Tools To Recognise Protein Surfaces and Probe Protein-Protein Interactions. *ChemMedChem* **2016**, *11* (8), 787–94. DOI: 10.1002/cmdc.201500450.
(3) Russo, A.; Aiello, C.; Grieco, P.; Marasco, D. Targeting "Undruggable" Proteins: Design of Synthetic Cyclopeptides. *Current Medicinal Chemistry* **2016**, *23* (8), 748–762. DOI: 10.2174/0929867323666160112122540.
(4) Josien, H.; Nair, A. G.; Ding, F.-X.; Guo, Y.; Chen, Y.-H.; Rao, A. U.; Liu, J.; Tong, L.; Sun, Z.; Lo, M. M. C.; et al. Discovery Process of Enlicitide, a Highly Engineered Macrocyclic Peptide Therapeutic, through Issue-Driven Fragment-Based Synthetic Assembly and SAR. *Journal of Medicinal Chemistry* **2026**, *69* (11), 13473–13491. DOI: 10.1021/acs.jmedchem.6c00463.
(5) Tanada, M.; Tamiya, M.; Matsuo, A.; Chiyoda, A.; Takano, K.; Ito, T.; Irie, M.; Kotake, T.; Takeyama, R.; Kawada, H.; et al. Development of Orally Bioavailable Peptides Targeting an Intracellular Protein: From a Hit to a Clinical KRAS Inhibitor. *Journal of the American Chemical Society* **2023**, *145* (30), 16610–16620. DOI: 10.1021/jacs.3c03886.
(6) Scola, P. M. Discovery of BMS-986238, a second-generation macrocyclic peptide inhibitor of programmed death-ligand 1 (PD-L1). In *ACS Spring 2025 National Meeting*, San Diego, CA, 2025/03/26, 2025; American Chemical Society.
(7) Fourie, A. M.; Cheng, X.; Chang, L.; Greving, C.; Li, X.; Knight, B.; Polidori, D.; Patrick, A.; Bains, T.; Steele, R.; et al. JNJ-77242113, a highly potent, selective peptide targeting the IL-23 receptor, provides robust IL-23 pathway inhibition upon oral dosing in rats and humans. *Scientific Reports* **2024**, *14* (1), 17515. DOI: 10.1038/s41598-024-67371-5.
(8) Scola, P. M. Discovery of BMS-986238, a second-generation macrocyclic peptide inhibitor of programmed death-ligand 1 (PD-L1). 2025.
(9) Johns, D. G.; Banka, P.; Zhou, Z.; Klapara, A.; Tsay, F.-R.; Kong, J.; Varsolona, R. J.; Desmond, R.; Maligres, P. E.; Marota, M.; et al. Compounds for treating conditions related to PCSK9 activity. WO2023023245A1, 2022.
(10) Dickinson, Q.; Meyer, J. G. Positional SHAP (PoSHAP) for Interpretation of machine learning models trained from biological sequences. *PLOS Computational Biology* **2022**, *18* (1), e1009736. DOI: 10.1371/journal.pcbi.1009736.
(11) Alleyne, C.; Amin, R. P.; Bhatt, B.; Bianchi, E.; Blain, J. C.; Boyer, N.; Branca, D.; Embrey, M. W.; Ha, S. N.; Jette, K.; et al. Series of Novel and Highly Potent Cyclic Peptide PCSK9 Inhibitors Derived from an mRNA Display Screen and Optimized via Structure-Based Design. *Journal of Medicinal Chemistry* **2020**, *63* (22), 13796–13824. DOI: 10.1021/acs.jmedchem.0c01084.
(12) Pissurlenkar, R. R. S.; Malde, A. K.; Khedkar, S. A.; Coutinho, E. C. Encoding Type and Position in Peptide QSAR: Application to Peptides Binding to Class I MHC Molecule HLA-A*0201. *QSAR & Combinatorial Science* **2007**, *26* (2), 189–203. DOI: 10.1002/qsar.200530184.
(13) Wan, F.; Kontogiorgos-Heintz, D.; de la Fuente-Nunez, C. Deep generative models for peptide design. *Digit Discov* **2022**, *1* (3), 195–208. DOI: 10.1039/d1dd00024a.
(14) Chen, S.; Lin, T.; Basu, R.; Ritchey, J.; Wang, S.; Luo, Y.; Li, X.; Pei, D.; Kara, L. B.; Cheng, X. Design of target specific peptide inhibitors using generative deep learning and molecular dynamics simulations. *Nature Communications* **2024**, *15* (1), 1611. DOI: 10.1038/s41467-024-45766-2.

(15) Das, P.; Sercu, T.; Wadhawan, K.; Padhi, I.; Gehrmann, S.; Cipcigan, F.; Chenthamarakshan, V.; Strobelt, H.; dos Santos, C.; Chen, P.-Y.; et al. Accelerated antimicrobial discovery via deep generative models and molecular dynamics simulations. *Nature Biomedical Engineering* **2021**, *5* (6), 613–623. DOI: 10.1038/s41551-021-00689-x.
(16) Surana, S.; Arora, P.; Singh, D.; Sahasrabuddhe, D.; Valadi, J. PandoraGAN: Generating Antiviral Peptides Using Generative Adversarial Network. *SN Computer Science* **2023**, *4* (5), 607. DOI: 10.1007/s42979-023-02203-3.
(17) Tang, S.; Zhang, Y.; Chatterjee, P. PepTune: De Novo Generation of Therapeutic Peptides with Multi-Objective-Guided Discrete Diffusion. 2024; p arXiv:2412.17780.
(18) Geylan, G.; Janet, J. P.; Tibo, A.; He, J.; Patronov, A.; Kabeshov, M.; Czechtizky, W.; David, F.; Engkvist, O.; De Maria, L. PepINVENT: generative peptide design beyond natural amino acids. *Chem Sci* **2025**, *16* (20), 8682–8696. DOI: 10.1039/d4sc07642g.
(19) Lee, S.; Kim, H. Bidirectional de novo peptide sequencing using a transformer model. *PLOS Computational Biology* **2024**, *20* (2), e1011892. DOI: 10.1371/journal.pcbi.1011892.
(20) Nie, Z.; Li, D.; Liu, Y.; Xu, F.; Zhang, H.; Huang, X.; Liu, X.; Wang, Z.; Ma, Y.; Ye, Y.; et al. A Unified Peptide Generative Framework via a Weakly Order-Dependent Autoregressive Language Model and Lifelong Learning. *Journal of Chemical Information and Modeling* **2025**, *65* (15), 7919–7935. DOI: 10.1021/acs.jcim.5c00623.
(21) Shah, A.; Guntuboina, C.; Barati Farimani, A. Peptide-GPT: Generative Design of Peptides using Generative Pre-trained Transformers and Bio-informatic Supervision. *arXiv e-prints* **2024**, arXiv:2410.19222. DOI: 10.48550/arXiv.2410.19222.
(22) Wang, R.; Zhang, H.; Nguyen, T.; Feng, S.; Pang, H.-W.; Yu, X.; Xiao, L.; Zhang, P. Z. PepThink-R1: LLM for Interpretable Cyclic Peptide Optimization with CoT SFT and Reinforcement Learning. *arXiv preprint arXiv:2508.14765* **2025**.
(23) Xu, X.; Xu, C.; He, W.; Wei, L.; Li, H.; Zhou, J.; Zhang, R.; Wang, Y.; Xiong, Y.; Gao, X. HELM-GPT: de novo macrocyclic peptide design using generative pre-trained transformer. *Bioinformatics* **2024**, *40* (6), btae364. DOI: 10.1093/bioinformatics/btae364.
(24) Rettie, S. A.; Campbell, K. V.; Bera, A. K.; Kang, A.; Kozlov, S.; De La Cruz, J.; Adebomi, V.; Zhou, G.; DiMaio, F.; Ovchinnikov, S.; et al. Cyclic peptide structure prediction and design using AlphaFold. *bioRxiv* **2023**, 2023.02.25.529956. DOI: 10.1101/2023.02.25.529956.
(25) Pacesa, M.; Nickel, L.; Schellhaas, C.; Schmidt, J.; Pyatova, E.; Kissling, L.; Barendse, P.; Choudhury, J.; Kapoor, S.; Alcaraz-Serna, A.; et al. One-shot design of functional protein binders with BindCraft. *Nature* **2025**, *646* (8084), 483–492. DOI: 10.1038/s41586-025-09429-6.
(26) Stark, H.; Faltings, F.; Choi, M.; Xie, Y.; Hur, E.; O'Donnell, T.; Bushuiev, A.; Uçar, T.; Passaro, S.; Mao, W.; et al. BoltzGen: Toward Universal Binder Design. 2025.
(27) Jumper, J.; Evans, R.; Pritzel, A.; Green, T.; Figurnov, M.; Ronneberger, O.; Tunyasuvunakool, K.; Bates, R.; Zidek, A.; Potapenko, A.; et al. Highly accurate protein structure prediction with AlphaFold. *Nature* **2021**, *596* (7873), 583–589. DOI: 10.1038/s41586-021-03819-2.
(28) Passaro, S.; Corso, G.; Wohlwend, J.; Reveiz, M.; Thaler, S.; Somnath, V. R.; Getz, N.; Portnoi, T.; Roy, J.; Stark, H.; et al. Boltz-2: Towards Accurate and Efficient Binding Affinity Prediction. *bioRxiv* **2025**, 2025.06.14.659707. DOI: 10.1101/2025.06.14.659707.
(29) Evans, R.; O'Neill, M.; Pritzel, A.; Antropova, N.; Senior, A.; Green, T.; Žídek, A.; Bates, R.; Blackwell, S.; Yim, J.; et al. Protein complex prediction with AlphaFold-Multimer. *bioRxiv* **2022**, 2021.10.04.463034. DOI: 10.1101/2021.10.04.463034.

(30) Siani, M. A.; Weininger, D.; Blaney, J. M. CHUCKLES: a method for representing and searching peptide and peptoid sequences on both monomer and atomic levels. *J Chem Inf Comput Sci* **1994**, *34* (3), 588–93. DOI: 10.1021/ci00019a017.
(31) Amarasinghe, K. N.; De Maria, L.; Tyrchan, C.; Eriksson, L. A.; Sadowski, J.; Petrović, D. a. Virtual Screening Expands the Non-Natural Amino Acid Palette for Peptide Optimization. *Journal of Chemical Information and Modeling* **2022**, *62* (12), 2999–3007. DOI: 10.1021/acs.jcim.2c00193.
(32) Devlin, J.; Chang, M.-W.; Lee, K.; Toutanova, K. BERT: Pre-training of Deep Bidirectional Transformers for Language Understanding. In *Proceedings of the 2019 Conference of the North American Chapter of the Association for Computational Linguistics: Human Language Technologies, Volume 1 (Long and Short Papers)*, Minneapolis, Minnesota, 2019/06/, 2019; Association for Computational Linguistics: pp 4171–4186. DOI: 10.18653/v1/N19-1423.
(33) Kotz, S.; Van Dorp, J. R. *Beyond beta: other continuous families of distributions with bounded support and applications*; World Scientific, 2004.
(34) Williams, R. J. Simple statistical gradient-following algorithms for connectionist reinforcement learning. *Machine Learning* **1992**, *8* (3-4), 229–256. DOI: 10.1007/BF00992696.
(35) Shao, Z.; Wang, P.; Zhu, Q.; Xu, R.; Song, J.; Zhang, M.; Li, Y. K.; Wu, Y.; Guo, D. DeepSeekMath: Pushing the Limits of Mathematical Reasoning in Open Language Models. *arXiv preprint arXiv:2402.03300* **2024**.
(36) Wildman, S. A.; Crippen, G. M. Prediction of Physicochemical Parameters by Atomic Contributions. *Journal of Chemical Information and Computer Sciences* **1999**, *39* (5), 868–873. DOI: 10.1021/ci990307l.
(37) *RDKit: Open-Source Cheminformatics*; 2024. https://www.rdkit.org.
(38) Pollard, V. W.; Malim, M. H. The HIV-1 rev protein. *Annual Review of Microbiology* **1998**, *52* (Volume 52, 1998), 491–532. DOI: 10.1146/annurev.micro.52.1.491.
(39) Zhuang, X.; Stahl, S. J.; Watts, N. R.; DiMattia, M. A.; Steven, A. C.; Wingfield, P. T. A Cell-penetrating Antibody Fragment against HIV-1 Rev Has High Antiviral Activity: characterization of the paratope *Journal of Biological Chemistry* **2014**, *289* (29), 20222–20233. DOI: 10.1074/jbc.M114.581090.
(40) Wu, H.; Mousseau, G.; Mediouni, S.; Valente, S. T.; Kodadek, T. Cell-Permeable Peptides Containing Cycloalanine Residues. *Angewandte Chemie International Edition* **2016**, *55* (41), 12637–12642. DOI: 10.1002/anie.201605745.
(41) Willett, P.; Barnard, J. M.; Downs, G. M. Chemical Similarity Searching. *Journal of Chemical Information Computer Sciences* **1998**, *38* (6), 983–996. DOI: 10.1021/ci9800211.
(42) McInnes, L.; Healy, J.; Saul, N.; Großberger, L. UMAP: Uniform Manifold Approximation and Projection. *Journal of Open Source Software* **2018**, *3* (29), 861. DOI: 10.21105/joss.00861.
(43) Rogers, D.; Hahn, M. Extended-Connectivity Fingerprints. *Journal of Chemical Information and Modeling* **2010**, *50* (5), 742–754. DOI: 10.1021/ci100050t.
(44) Yu, J.; Zheng, Y.; Koh, H. Y.; Pan, S.; Wang, T.; Wang, H. Collaborative Expert LLMs Guided Multi-Objective Molecular Optimization. 2025; p arXiv:2503.03503.
(45) Kim, H.; Jang, Y.; Ahn, S. MT-Mol: Multi Agent System with Tool-based Reasoning for Molecular Optimization. In *Findings of the Association for Computational Linguistics: EMNLP 2025*, 2025; Association for Computational Linguistics.

## Supplementary Information

**PepEVOLVE: Position-Aware Dynamic Peptide Optimization via Group-Relative Advantage**

Trieu Nguyen[1,3], Peter Zhiping Zhang[1], Nicolas Boyer[1], Cheng Fang[2], Liying Zhang[1], Sebastian Schneider[1], Hao-Wei Pang[2*], Shasha Feng[1*]

1. Merck & Co., Inc., Boston, MA, USA, 02115
2. Merck & Co., Inc., Cambridge, MA, USA, 02141
3. University of South Florida, Tampa, FL, USA, 33620

*Corresponding authors (hao-wei.pang@merck.com, shasha.feng@merck.com)

**Table of Contents**

## S1. Example of Masking Strategies

To illustrate how the masking strategies described in Section 2.5 Evolving Strategy work in practice, consider the peptide sequence

$$p = "r_1|r_2|r_3|r_4|r_5",$$

where the positions selected for optimization are $O = \{2,3,4\}$, giving $| O |= 3$input contexts.

### Single-agent Mode

Under the self-mask strategy, each context masks only the position being optimized:

$$\text{agent} \leftarrow "r_1|?|r_3|r_4|r_5", \text{agent} \leftarrow "r_1|r_2|?|r_4|r_5", \text{agent} \leftarrow "r_1|r_2|r_3|?|r_5".$$

In contrast, the neighbor-mask strategy masks neighboring positions around the target position, producing:

$$\text{agent} \leftarrow "r_1|r_2|?|?|r_5", \text{agent} \leftarrow "r_1|?|r_3|?|r_5", \text{agent} \leftarrow "r_1|?|?|r_4|r_5".$$

### Multi-agent Mode

Under the self-mask strategy, each context masks only the position being optimized and each context is handled by a corresponding agent:

$$\text{agent}_2 \leftarrow "r_1|?|r_3|r_4|r_5", \text{agent}_3 \leftarrow "r_1|r_2|?|r_4|r_5", \text{agent}_4 \leftarrow "r_1|r_2|r_3|?|r_5".$$

In contrast, the neighbor-mask strategy masks neighboring positions around the target position and each context is handled by a corresponding agent:

$$\text{agent}_2 \leftarrow "r_1|r_2|?|?|r_5", \text{agent}_3 \leftarrow "r_1|?|r_3|?|r_5", \text{agent}_4 \leftarrow "r_1|?|?|r_4|r_5".$$

## S2. Score Analysis and Model Collapse Behavior Observed

It is important to note that in **Figure 7**, the mean score is computed as the sum of all unique peptide scores divided by the number of unique peptides, consistent with the original PepINVENT method. This approach is valid only when the uniqueness of generated peptides remains close to 100%. In certain PepEVOLVE configurations, we observed a decline in peptide uniqueness over training steps. To address this, we conducted a refined analysis of score progression by calculating the mean score based on all generated peptides, rather than only the unique ones (**Figure S1**).

By doing this, we observe pronounced fluctuations after approximately 400 steps in both cases involving self-mask context visibility as illustrated in **Figure 7**. We interpret this as evidence of model collapse, which appears to arise from two interacting factors. First, the group-relative advantage mechanism tends to overemphasize certain score components, effectively over-squashing the optimization signal. Second, within the self-mask context, the model generates only a single monomer at a time. This setting involves significantly fewer tokens than the neighbor-mask context, which requires roughly three times as many tokens for generation. The limited token context amplifies the over-squashing effect, reducing learning diversity and ultimately leading to model collapse.

Moreover, because group-relative advantage normalizes rewards across all peptides in a group, even high-quality peptides may be penalized if others in the group achieve better scores. Over time, this causes the model to converge toward a narrow subset of “top-tier” peptides, reducing variability and leading to the generation of fewer unique sequences.

The observed fluctuations between very high and very low scores can be further explained by differences in the optimization landscapes across input contexts. Some contexts experience collapse, while others remain stable. In collapsed contexts, the model produces only a small number of unique peptides, resulting in lower aggregate scores, not because the generated peptides themselves are of poor quality, but because the score normalization is based on the total number of generated samples.

This relationship is further illustrated in **Figure S10**. The self-mask context produces fewer unique peptides overall, but among the peptides with scores between 0.9 and 1.0, it actually performs better than all other contexts. This matches our earlier explanation: as the model collapses, only the very best peptides remain.

Nevertheless, all four configurations of PepEVOLVE still outperform PepINVENT in producing optimized peptides. Notably, PepINVENT fails to generate any peptides with a score above 0.8, whereas PepEVOLVE generates a significantly higher portion of peptides with score above 0.8.

### S3. Ablation Study of PepEVOLVE Components

To ensure the robustness of our findings, we conducted an ablation study evaluating the individual and combined contributions of PepEVOLVE's core components against the PepINVENT baseline. The three primary algorithmic additions evaluated were CHUCKLES shifting during the pretraining phase, alongside Top-K extraction and group-relative advantage (GRA) during the reinforcement learning (RL) optimization phase. Because GRA mathematically relies on the grouped candidate outputs generated by the Top-K extraction process, it cannot be evaluated as an isolated mechanism. Consequently, we assessed the following configurations with dynamic mask always on: CHUCKLES Shift alone, Top-K alone, Top-K combined with CHUCKLES Shift, Top-K combined with GRA, and the complete PepEVOLVE model (CHUCKLES Shift + Top-K + GRA).

As illustrated in **Figure Figure** , the learning curves across the evaluated agent configurations (including NM, NS, SM, and SS) demonstrate that each integrated component plays an important role in enhancing optimization performance. The baseline PepINVENT model and the CHUCKLES Shift-only configuration consistently yield the lowest mean total scores. The introduction of Top-K extraction provides a distinct performance boost. However, the most significant performance improvements are observed when Top-K extraction is paired with GRA. Top-K + GRA rapidly drives the mean total score toward 0.9, illustrating that GRA can effectively stabilize policy updates and successfully steers the model toward improved sequence spaces. Ultimately, the full PepEVOLVE framework, which combines all three components, consistently achieves the highest and most stable mean total scores across the training steps.

To further illustrate the impact of these components, **Figure S** shows the score density distributions of the generated peptides across different configurations. The baseline PepINVENT and the CHUCKLES Shift-only models produce peptides with scores predominantly concentrated around 0.6. Incorporating the Top-K extraction mechanism visibly broadens this distribution, shifting a fraction of the generated sequences into higher-scoring regions. However, the most pronounced improvement is observed when Group Relative Advantage (GRA) is active. Both the Top-K + GRA and the full PepEVOLVE configurations dramatically shift the bulk of the score mass to form a sharp peak near 0.9. This consistent rightward shift across all agent configurations shows that combining Top-K with GRA is critical for maximizing the yield of top-tier peptide candidates.

The quantitative impact of these components is further detailed in Table 1 and Table 2, which summarize the scores of the top 1 and top 10 generated peptides, respectively. The baseline PepINVENT model achieved a top 1 score of 0.8862 and a top 10 average of 0.8636. Incorporating only the CHUCKLES Shift provided a marginal increase, yielding a top 1 score of 0.9024 and a top 10 average of 0.8933. The addition of Top-K extraction demonstrated noticeable improvements across all agent configurations (NM, NS, SM, SS), pushing top 1 scores into the 0.9190–0.9696 range. However, the synergetic effect of combining Top-K with GRA proved most impactful; for instance, the Top-K + GRA configuration achieved top 1 scores up to 0.9798 (under the SS mode) and maintained a remarkably high top 10 average of 0.9752 ± 0.0020. Ultimately, the complete PepEVOLVE model (CHUCKLES Shift + Top-K + GRA) demonstrated the most robust performance, consistently generating elite candidates with top 10 averages exceeding 0.9577 across all modes. This confirms that these combined mechanisms reliably produce a high volume of quality peptides rather than isolated statistical outliers.

The ablation study clearly illustrates that while individual components like CHUCKLES shifting and Top-K extraction offer necessary incremental benefits, it is the integration of Group Relative Advantage (GRA) that unlocks the most substantial gains in peptide optimization. Together, these elements form a robust architecture in PepEVOLVE that consistently outperforms the baseline, ensuring both rapid convergence to optimal chemical spaces and a reliable, high-quality yield of peptides.

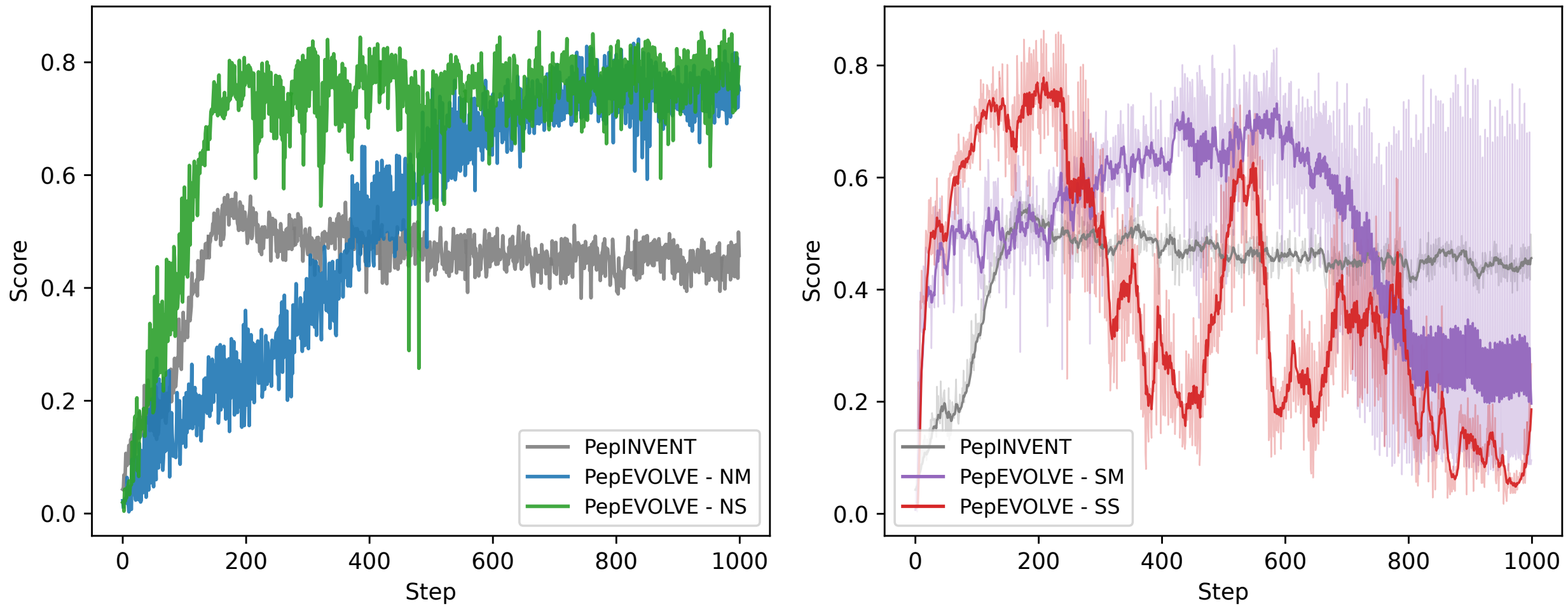

**Figure S10.** Generated peptide scores over the steps when the mean score is calculated using all generated peptides rather than only unique ones. Left: Comparison between PepINVENT and PepEVOLVE (neighbor-mask) configurations. Right: Comparison between PepINVENT and PepEVOLVE (self-mask) configurations, shown with exponential moving average smoothing.

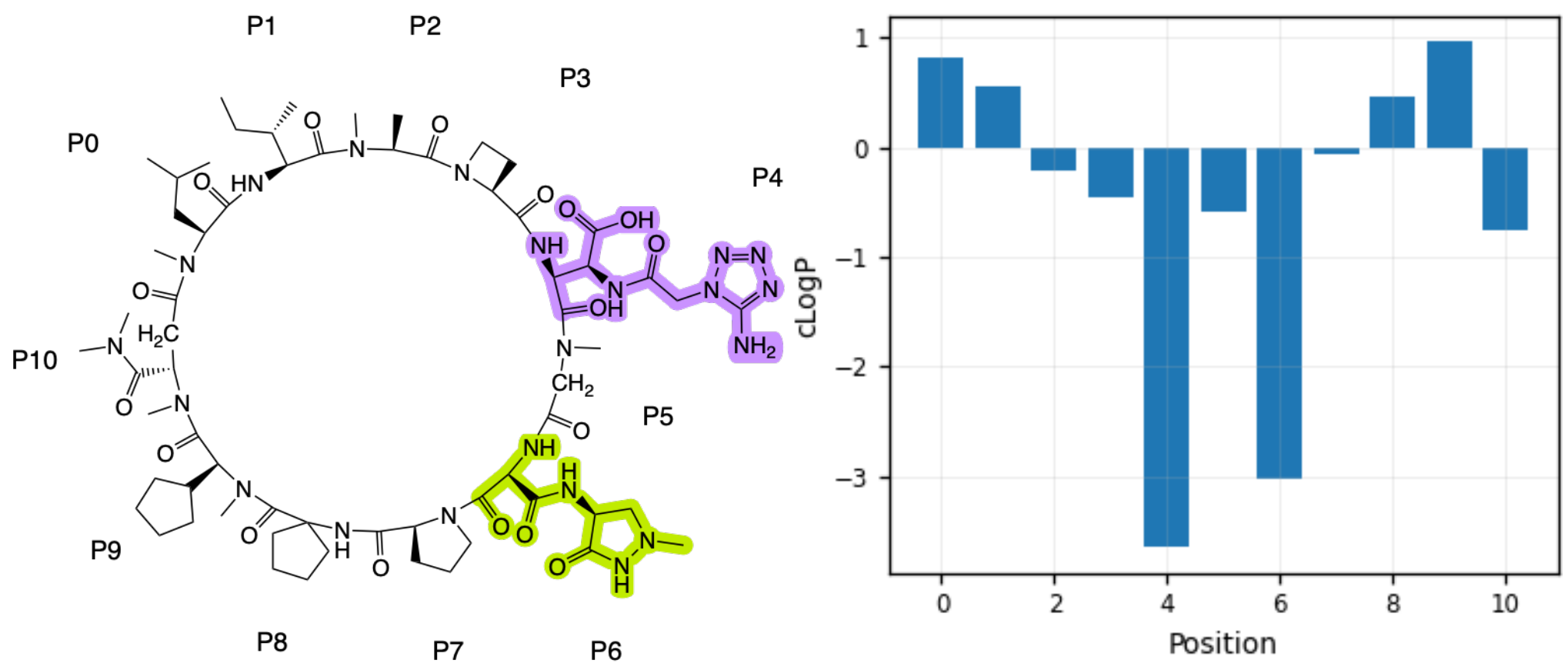

**Figure S2.** Starting peptide used for router policy convergence in the adversarial test. Left: The peptide extracted from the LUNA18 test case. Right: Per-residue cLogP using amino acid at each position. cLogP calculated using Crippen method implemented in RDKit. CHUCKLES for adversarial router test is

"N3(C)[C@@H](CC(C)C)C(=O)|N[C@@H]([C@H](CC)C)C(=O)|N(C)[C@@H](C)C(=O)|N1[C@@H](CC1)C(=O)|N[C@@H]([C@H](NC(=O)Cn1c(N)nnn1)C(=O)O)C(=O)|N(C)CC(=O)|N[C@@H](C(=O)N[C@H]1CN(C)NC1=O)C(=O)|N1[C@@H](CCC1)C(=O)|NC1(CCCC1)C(=O)|N(C)[C@@H](C1CCCC1)C(=O)|N(C)[C@H](C(=O)N(C)C)CC3(=O)".

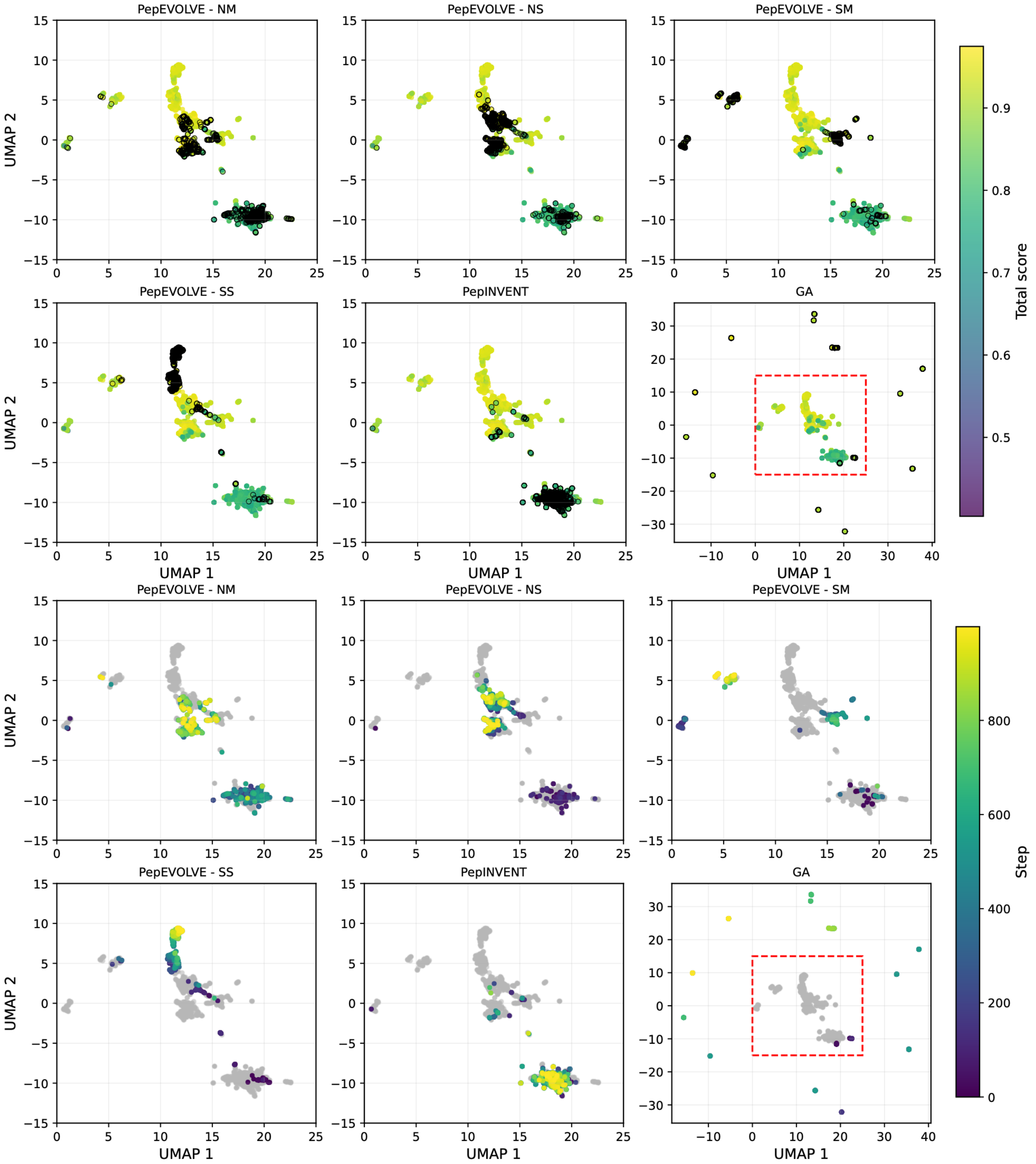

**Figure S3.** Chemical space analysis for PepEVOLVE, PepINVENT, and GA on Rev-binding macrocycle optimization. Top: UMAP projection of chemical space colored by total score. Molecules generated by each approach are circled. Bottom: UMAP projection of chemical space colored by steps. Featurization was performed using count-based Morgan fingerprints. Non-GA methods are zoomed in; the zoomed-in region is indicated by a dashed box in the GA panel.

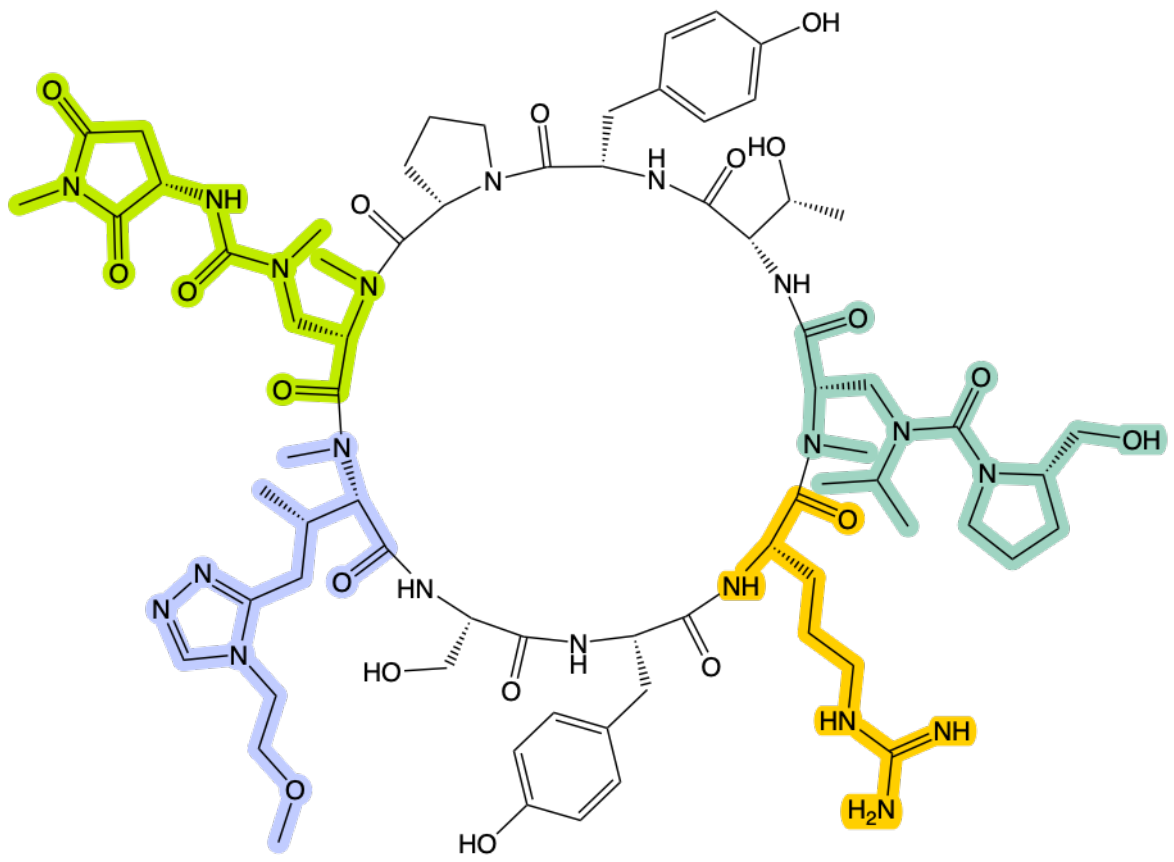

**Figure S4.** Top-scoring molecule from GA run for RBP case.

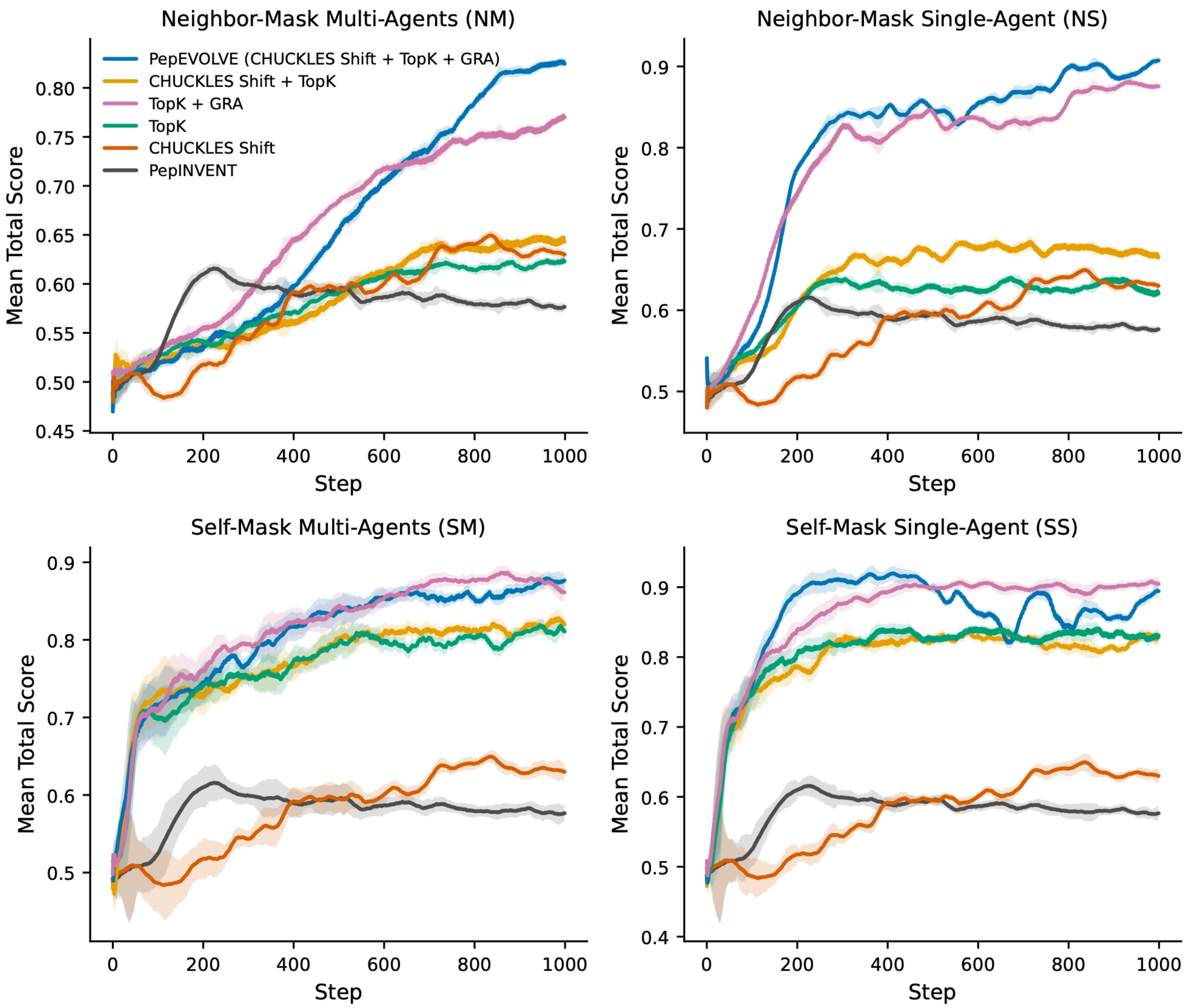

**Figure S5.** Learning curves demonstrating the optimization performance of various ablation configurations under each masking-agent strategy. Color coding is shared among the four panels.

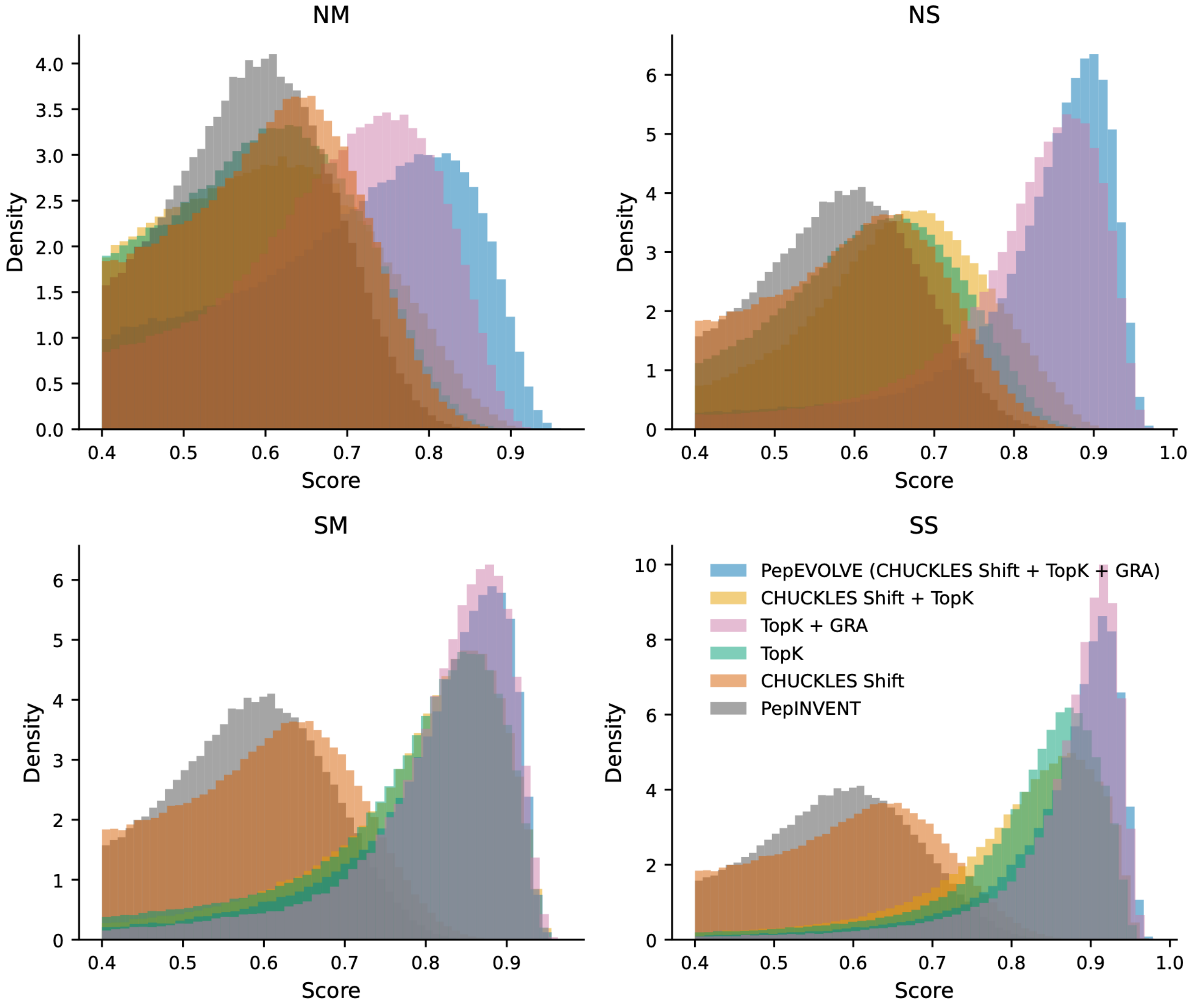

**Figure S6.** Score density distributions of generated peptides across different ablation configurations under each masking-agent strategy. Color coding is shared among the four panels.

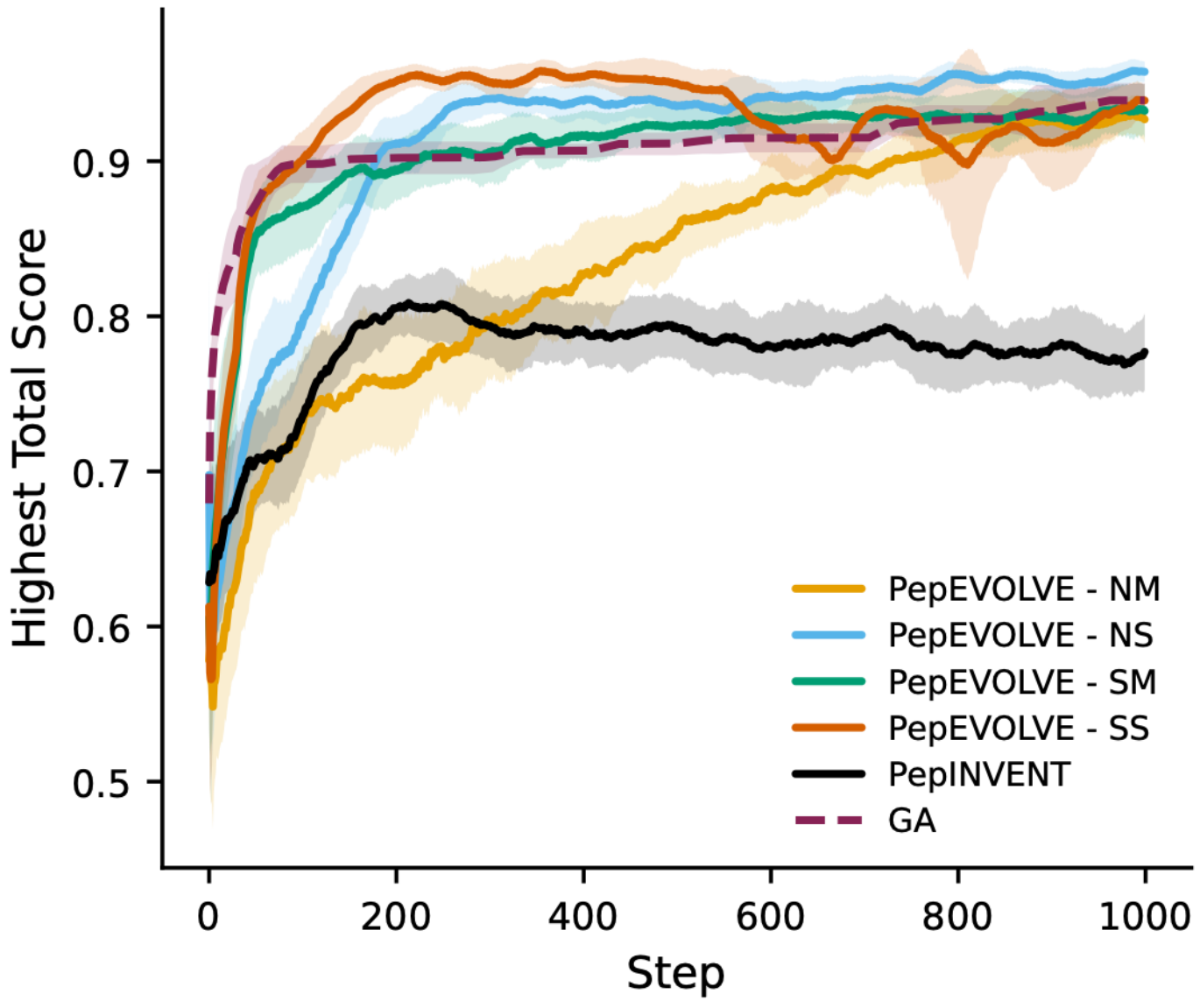

**Figure S7.** Highest total score of generated peptides from each optimization step on Rev-binding macrocycle optimization under a permeability-aware multi-objective.

**Table S1.** SMARTS alert patterns for RBP optimization.

| SMARTS Pattern | Common Functional Group / Context |
|---|---|
| [#8][#8] | Peroxides (highly reactive, potential explosive hazard, or easily degraded). |
| [#6;+] | Carbocations (extremely reactive intermediates, usually unstable in screening libraries). |
| [#7;!n][S;!$(S(=O)=O)] | Sulfenamides or unstable nitrogen-sulfur single bonds (often chemically reactive/labile). |
| [#7;!n][#7;!n] | Hydrazines or related acyclic $N-N$ single bonds (often associated with toxicity or instability). |
| C#C | Alkynes (acetylenic groups, can sometimes be reactive or act as metabolic liabilities). |
| C(=[O,S])[O,S] | Carboxylic acids, esters, thioesters, or carbamothioates. |
| [#7;!n][C;!$(C(=[O,N])[N,O])][#16;!s] | Hemiaminals / Aminomethyl thioethers (potentially hydrolytically unstable linkages). |
| [#7;!n][C;!$(C(=[O,N])[N,O])][#7;!n] | Aminals (often unstable and prone to hydrolysis into aldehydes/ketones and amines). |
| [#7;!n][C;!$(C(=[O,N])[N,O])][#8;!o] | Hemiaminals or their ethers (unstable/reversible linkages). |
| [#8;!o][C;!$(C(=[O,N])[N,O])][#16;!s] | Monothioacetals or monothioketals (hydrolytically labile). |
| [#8;!o][C;!$(C(=[O,N])[N,O])][#8;!o] | Acetals or Ketals (acid-labile protecting groups or unstable hemiacetals). |
| [#16;!s] [C;!$(C(=[O,N])[N,O])] [#16;!s] | Dithioacetals or dithioketals. |

**Table S2.** Transformation and desirability functions for RBP optimization.

| Scoring Component | Weight | Desirability Transformation | Specific Parameters / Thresholds |
|---|---|---|---|
| Permeability model | 3 | No transformation | Uses permeability model published by PepINVENT. |
| Maximum ring size | 1 | Double sigmoid* | $low = 14, high = 30, \beta = 10, k_1 = 20, k_2 = 20$ |
| Lipophilicity | 1 | Reverse sigmoid** | $low = -5, high = 0, k = 0.5$ |
| Custom alerts | 1 | Step function | SMARTS-based structural alerts. See Table S1 for details. |

* Double sigmoid function is defined as: $f(x) = \frac{1}{1+10^\wedge\left(-k_1\frac{x-low}{\beta}\right)} - \frac{1}{1+10^\wedge\left(-k_2\frac{x-high}{\beta}\right)}$

** Reverse sigmoid function is defined as $f(x) = \frac{1}{1+10^\wedge\left(k(x-\frac{high+low}{2})\times\frac{10}{high-low}\right)}$

**Table S3**. Highest score of the peptide achieved across the PepINVENT baseline and various ablation configurations. Evaluated under the four PepEVOLVE modes (neighbor-mask multi-agent [NM], neighbor-mask single-agent [NS], self-mask multi-agent [SM], and self-mask single-agent [SS]). The scoring performance of the best two configurations are annotated in bold.

| | NM | NS | SM | SS |
|---|---|---|---|---|
| PepINVENT | | 0.886 | | |
| CHUCKLES Shift | | 0.902 | | |
| TopK | 0.932 | 0.919 | 0.963 | 0.970 |
| TopK + GRA | 0.935 | **0.975** | 0.963 | **0.980** |
| CHUCKLES Shift + TopK | **0.938** | 0.938 | **0.967** | 0.973 |
| PepEVOLVE | **0.961** | **0.975** | **0.967** | **0.979** |

**Table S4**. Average scores and standard deviations of the Top 10 generated peptides across the PepINVENT baseline and various ablation configurations. Evaluated under the four PepEVOLVE modes (neighbor-mask multi-agent [NM], neighbor-mask single-agent [NS], self-mask multi-agent [SM], and self-mask single-agent [SS]). The scoring performance of the best two configurations are annotated in bold.

| | NM | NS | SM | SS |
|---|---|---|---|---|
| PepINVENT | | 0.864 | | |
| CHUCKLES Shift | | 0.893 | | |
| TopK | 0.914 ± 0.009 | 0.902 ± 0.008 | 0.959 ± 0.002 | 0.966 ± 0.002 |
| TopK + GRA | 0.929 ± 0.003 | **0.973 ± 0.002** | 0.960 ± 0.001 | **0.975 ± 0.002** |
| CHUCKLES Shift + TopK | **0.930 ± 0.006** | 0.930 ± 0.004 | **0.963 ± 0.002** | 0.970 ± 0.002 |
| PepEVOLVE | **0.958 ± 0.002** | **0.973 ± 0.001** | **0.963 ± 0.003** | **0.976 ± 0.001** |